%% file: main.tex
\documentclass[sigconf]{acmart} 
\input{notation}

\usepackage{algorithm}
\usepackage{algpseudocode}

\AtBeginDocument{%
  \providecommand\BibTeX{{%
    \normalfont B\kern-0.5em{\scshape i\kern-0.25em b}\kern-0.8em\TeX}}}

\copyrightyear{2021}
\acmYear{2021}
\setcopyright{acmcopyright}\acmConference[RecSys '21]{Fifteenth ACM Conference on Recommender Systems}{September 27-October 1, 2021}{Amsterdam, Netherlands}
\acmBooktitle{Fifteenth ACM Conference on Recommender Systems (RecSys '21), September 27-October 1, 2021, Amsterdam, Netherlands}
\acmPrice{15.00}
\acmDOI{10.1145/3460231.3474245}
\acmISBN{978-1-4503-8458-2/21/09}

\begin{document}
\fancyhead{}
\title{Evaluating the Robustness of Off-Policy Evaluation}


\author{Yuta Saito}
\authornote{Both authors contributed equally to this work.}
\affiliation{
\institution{Hanjuku-Kaso Co., Ltd.}
\country{Tokyo, Japan}
}
\email{saito@hanjuku-kaso.com}

\author{Takuma Udagawa}
\authornotemark[1]
\affiliation{
\institution{Sony Group Corporation}
\country{Tokyo, Japan}
}
\email{Takuma.Udagawa@sony.com}

\author{Haruka Kiyohara}
\authornote{This work was done during their internship at Hanjuku-Kaso Co., Ltd.}
\affiliation{
\institution{Tokyo Institute of Technology}
\country{Tokyo, Japan}
}
\email{kiyohara.h.aa@m.titech.ac.jp}

\author{Kazuki Mogi}
\authornotemark[2]
\affiliation{
\institution{Stanford University}
\country{California, United States}
}
\email{kmogi@stanford.edu}

\author{Yusuke Narita}
\affiliation{
\institution{Yale University}
\country{Connecticut, United States}
}
\email{yusuke.narita@yale.edu}

\author{Kei Tateno}
\affiliation{
\institution{Sony Group Corporation}
\country{Tokyo, Japan}
}
\email{Kei.Tateno@sony.com}

\renewcommand{\shortauthors}{Saito. et al}

\begin{abstract}
\textit{Off-policy Evaluation} (OPE), or offline evaluation in general, evaluates the performance of hypothetical policies leveraging only offline log data. It is particularly useful in applications where the online interaction involves high stakes and expensive setting such as precision medicine and recommender systems. Since many OPE estimators have been proposed and some of them have hyperparameters to be tuned, there is an emerging challenge for practitioners to select and tune OPE estimators for their specific application. Unfortunately, identifying a reliable estimator from results reported in research papers is often difficult because the current experimental procedure evaluates and compares the estimators' performance on a narrow set of hyperparameters and evaluation policies. Therefore, it is difficult to know which estimator is safe and reliable to use. In this work, we develop \textit{Interpretable Evaluation for Offline Evaluation} (IEOE), an experimental procedure to evaluate OPE estimators' robustness to changes in hyperparameters and/or evaluation policies in an interpretable manner. Then, using the IEOE procedure, we perform extensive evaluation of a wide variety of existing estimators on Open Bandit Dataset, a large-scale public real-world dataset for OPE. We demonstrate that our procedure can evaluate the estimators' robustness to the hyperparamter choice, helping us avoid using unsafe estimators. Finally, we apply IEOE to real-world e-commerce platform data and demonstrate how to use our protocol in practice.
\end{abstract}



\maketitle

\section{Introduction}
Interactive bandit and reinforcement learning algorithms have been used to optimize decision making in many real-life scenarios such as precision medicine, recommender systems, advertising, etc.
We often use these algorithms to maximize the expected reward, but they also produce log data valuable for evaluating and redesigning future decision making. 
For example, the logs of a news recommender system record which news article was presented and whether the user read it, giving the decision maker a chance to make its recommendation more relevant. 
Exploiting log data is, however, more difficult than conventional supervised machine learning. 
This is because the result is only observed for the action chosen by the algorithm but not for all the other actions the system could have taken. 
The logs are also biased, as the logs overrepresent the actions favored by the algorithm used to collect the data. 
Online experiment or A/B test is a potential solution to this issue.
It compares the performance of counterfactual algorithms in an online environment, enabling unbiased evaluation and comparison.
However, A/B testing counterfactual algorithms is often difficult, since deploying a new policy to a real environment is time-consuming and may damage user satisfaction~\citep{gilotte2018offline,saito2020doubly}.

This motivates us to study \textit{Off-policy Evaluation} (OPE), which aims to estimate the performance of an evaluation policy using only log data collected by a behavior policy.
Such an evaluation allows us to compare the performance of candidate policies safely and helps us decide which policy to deploy in the field. 
This alternative offline evaluation approach thus has the potential to overcome the above issues with the online A/B test approach.

With growing interest in OPE, the research community has produced a number of estimators, including Direct Method (DM)~\citep{beygelzimer2009offset}, Inverse Probability Weighting (IPW)~\citep{precup2000eligibility,strehl2010learning}, Self-Normalized IPW (SNIPW)~\citep{swaminathan2015self}, Doubly Robust (DR)~\citep{dudik2014doubly}, Switch-DR~\citep{wang2017optimal}, and Doubly Robust with Optimistic Shrinkage (DRos)~\citep{su2020doubly}. 

One emerging challenge with this trend is that there is a need for practitioners to select and tune appropriate hyperparameters for OPE estimators for their specific application~\citep{su2020adaptive,voloshin2019empirical}.
For example, DM first estimates the expected reward function using an arbitrary machine learning method, then uses its estimate for OPE. 
Therefore, one has to identify a good machine learning method to estimate the expected reward before the offline evaluation phase.
Identifying the appropriate machine learning method for DM is difficult, because its accuracy cannot be easily quantified from bandit data~\citep{jiang2016doubly}.
Sophisticated estimators such as Switch-DR~\citep{wang2017optimal} and DRos~\citep{su2020doubly} show improved offline evaluation performance in some experiments.
However, these estimators have a larger number of hyperparameters to be tuned compared to the baseline estimators. 
A difficulty here is that the estimation accuracy of OPE estimators is highly sensitive to the choice of hyperparameters, as implied in empirical studies~\citep{voloshin2019empirical,saito2020open}.
When we rely on OPE in real-world applications, it is desirable to use an estimator that is robust to the choice of hyperparameters and achieves accurate evaluations without requiring significant hyperparameter tuning.
Moreover, we want the estimators to be robust to other possible configuration changes such as evaluation policies.
An estimator of this type is preferable, because tuning hyperparameters of OPE estimators with only logged bandit data is challenging in nature, and we often apply an estimator to several different policies to compare the performance of candidate policies offline. 
The aim of this paper is thus to enable a safer OPE practice by developing a procedure to evaluate the estimators' robustness.
\\

\textbf{Current dominant evaluation procedures.}
The current evaluation procedure used in OPE research is not suitable for evaluating the estimators' robustness.
Almost all OPE papers evaluate the estimator's performance for a single given set of hyperparameters and an arbitrary evaluation policy~\citep{saito2020open,dudik2014doubly,wang2017optimal,su2019cab,su2020doubly,narita2019efficient,vlassis2019design,liu2019triply,farajtabar2018more,kato2020off}.
Even though it is common to iterate trials with different random seeds to provide an estimate of the performance, this procedure cannot evaluate the estimators' robustness to hyperparameter choices or the changes in evaluation policies, which is critical in real-world scenarios.
The estimator's performance derived from this common procedure does not properly account for the uncertainty in offline evaluation performance, as the reported performance metric is a single random variable drawn from the distribution over the estimator's performance.
Consequently, choosing an appropriate OPE estimator is difficult, as their robustness to hyperparameter choices or the changes in evaluation policies are not quantified in existing experiments.
\\

\begin{figure}
    \centering
    \includegraphics[width=0.8\linewidth]{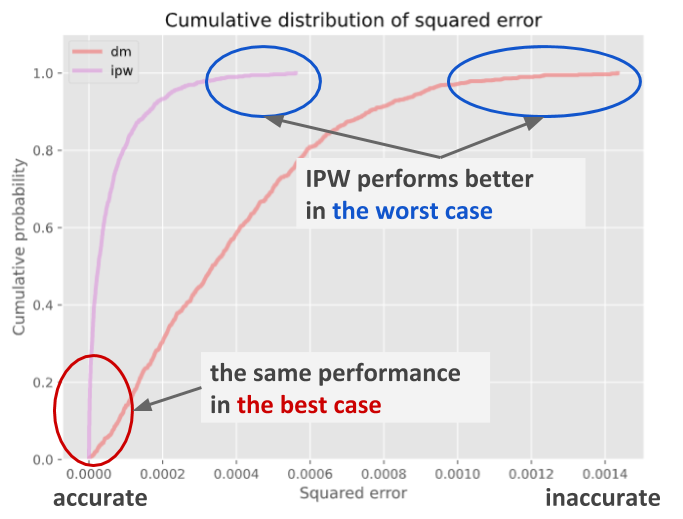}
    \caption{An example output of the proposed evaluation procedure for offline evaluation}
    \label{fig:example_cdf_results}
\end{figure}

\textbf{Contributions.}
Motivated towards promoting a reliable use of OPE in practice, we develop an interpretable and scalable evaluation procedure for OPE estimators that quantifies their robustness to the choice of hyperparameters and possible changes in evaluation policies.
Our evaluation procedure compares several OPE estimators as depicted in Figure~\ref{fig:example_cdf_results}.
This figure compares the offline evaluation performance of IPW and DM by illustrating their accuracy distributions as we vary their hyperparameters, evaluation policies, and random seeds.
The x-axis is the squared error in offline evaluation; a lower value indicates that an estimator is more accurate.
The figure is visually interpretable, and in this case, we are confident that IPW is better, having lower squared errors with high probability, being robust to the changes in configurations, and being more accurate even in the worst case.
In addition to developing the evaluation procedure, we have implemented open-source Python software, \textbf{\textsl{pyIEOE}}\footnote{https://github.com/sony/pyIEOE}, so that researchers can easily implement our procedure in their experiments, and practitioners can identify the best estimator for their specific environment.

Using our procedure and software, we evaluate a wide variety of existing OPE estimators on Open Bandit Dataset~\citep{saito2020open} (Section~\ref{sec:obd}) and several classification datasets (Appendix~\ref{app:benchmark}). Through these extensive experiments, we demonstrate that IEOE can provide informative results, in particular the estimators' robustness to the hyperparameter settings and evaluation policy changes, which could not be obtained using typical experimental procedure in OPE research.

Finally, as a proof of concept, we use our procedure to select the best estimator for the offline evaluation of coupon treatment policies on a real-world e-commerce platform.
The platform uses OPE to improve its coupon optimization policy safely without implementing A/B tests.
However, the platform's data scientists do not know which OPE estimator is appropriate for their setting.
We apply our procedure to provide an appropriate estimator choice for the platform.
This real-world application demonstrates how to use our procedure to reduce uncertainty and risk that we face in real-world offline evaluation.

Our contributions are summarized as follows.
\begin{itemize}
    \item We develop an experimental procedure called IEOE that is useful for identifying robust estimators and avoid the use of estimators sensitive to configuration changes.
    \item We have implemented \textbf{\textsl{pyIEOE}}, open-source Python software, that facilitates the use of our experimental procedure both in research and in practice.
    \item We conduct comprehensive benchmark experiments on public datasets and demonstrate that IEOE is useful for identifying estimators sensitive to configuration changes, and thus can help avoid potential failures in OPE.
    \item We apply IEOE to a real-world OPE application and demonstrate how this procedure helps us safely conduct OPE in practice.
\end{itemize}

\section{Off-Policy Evaluation} \label{sec:ope}

\subsection{Setup}\label{sec:setup}
We consider a general contextual bandit setting.
Let $r \in [0, r_{\mathrm{max}}]$ denote a reward or outcome variable (e.g., whether a coupon assignment results in an increase in revenue) and $a \in \calA$ be a discrete action. 
We let $x \in \calX$ be a context vector (e.g., the user's demographic profile) that the decision maker observes when picking an action. 
Rewards and contexts are sampled from unknown probability distributions $p (r \mid x, a)$ and $p(x)$, respectively.
We call a function $\pi: \calX \rightarrow \Delta(\calA)$ a \textit{policy}.
It maps each context $x \in \calX$ into a distribution over actions, where $\pi (a \mid x)$ is the probability of taking action $a$ given context vector $x$.

Let $\calD := \{(x_i,a_i,r_i)\}_{i=1}^n$ be a historical logged bandit feedback with $n$ observations.
$a_i$ is a discrete variable indicating which action in $\mathcal{A}$ is chosen for individual $i$. 
$r_i$ and $x_i$ denote the reward and the context observed for individual $i$.
We assume that a logged bandit feedback dataset is generated by a \textit{behavior policy} $\pi_b$ as follows:
\begin{align*}
  \{(x_i,a_i,r_i)\}_{i=1}^n \ \sim \ \prod_{i=1}^n p(x_i) \pi_b (a_i \mid x_i) p(r_i \mid x_i, a_i),
\end{align*}
where each context-action-reward triplet is sampled independently from the identical product distribution.
Then, for a function $f(x,a,r)$, we use $\mE_{n} [f] := n^{-1} \sum_{(x_i, a_i, r_i) \in \calD} f(x_i, a_i, r_i)$ to denote its empirical expectation over $n$ observations in $\calD$.
We also use $q(x,a) := \mE_{r \sim p(r|x,a)} [ r \mid x, a ]$ to denote the mean reward function for a given context and action.

In OPE, we are interested in using historical logged bandit data to estimate the following \textit{policy value} of a given \textit{evaluation policy} $\pi_e$ which might be different from $\pi_b$: 
\begin{align*}
    \trueV := \mE_{(x,a,r) \sim p(x) \pi_e (a \mid x) p(r \mid x, a)} [r] .
\end{align*}
Estimating $\trueV$ before deploying $\pi_e$ in an online environment is useful in practice, because $\pi_e$ may perform poorly.
Additionally, this makes it possible to select an evaluation policy that maximizes the policy value by comparing their estimated performances without incurring additional implementation cost.

\subsection{Existing OPE Estimators}\label{sec:estimators}
Given the policy value as the estimand, the goal of researchers is to propose an accurate estimator.
OPE estimator $\hat{V}$ estimates the policy value of an arbitrary evaluation policy as $\trueV \approx \hat{V} (\pi_e; \calD, \theta)$, where $\calD$ is an available logged bandit feedback dataset, and $\theta$ is a set of pre-defined \textit{hyperparameters} of $\hat{V}$.

Below, we summarize the definitions and properties of several existing OPE estimators.
We also summarize their built-in hyperparameters in Table~\ref{tab:hyperparams}.

\begin{table*}[t]
\large
\centering
\caption{Hyperparameters of the OPE estimators}
\def\arraystretch{1.2}
\scalebox{0.80}{
\begin{tabular}{c|cc}
\toprule
\textbf{OPE Estimators} && \textbf{Hyperparameters}  \\\midrule \midrule
Direct Method (DM) && $\hat{q}$, $K$ \\
Inverse Probability Weighting with Pessimistic Shrinkage (IPWps)~\citep{su2020doubly,strehl2010learning} && $\lambda$, ($\hat{\pi}_b$) \\
Self-Normalized Inverse Probability Weighting (SNIPW)~\citep{swaminathan2015self} && ($\hat{\pi}_b$) \\
Doubly Robust with Pessimistic Shrinkage (DRps)~\citep{dudik2014doubly,su2020doubly} && $\hat{q}$, $K$, $\lambda$, ($\hat{\pi}_b$) \\
Self-Normalized Doubly Robust (SNDR) && $\hat{q}$, $K$, ($\hat{\pi}_b$) \\
Switch Doubly Robust (Switch-DR)~\citep{wang2017optimal} && $\hat{q}$, $K$, $\tau$, ($\hat{\pi}_b$) \\
Doubly Robust with Optimistic Shrinkage (DRos)~\citep{su2020doubly} && $\hat{q}$, $K$, $\lambda$, ($\hat{\pi}_b$) \\
\bottomrule
\end{tabular}
}
\vskip 0.1in
\raggedright
\fontsize{8.5pt}{8.5pt}\selectfont \textit{Note}:
$\hat{q}$ is an estimator for the mean reward function constructed by an arbitrary machine learning method.
$K$ is the number of folds in the cross-fitting procedure.
$\hat{\pi}_b$ is an estimated behavior policy.
This is unnecessary when we know the true behavior policy, and thus it is in parentheses.
$\tau$ and $\lambda$ are non-negative hyperparameters for defining the corresponding estimators.
\label{tab:hyperparams}
\end{table*}

\paragraph{Direct Method (DM)}
DM~\citep{beygelzimer2009offset} first trains a supervised machine learning method, such as ridge regression, to estimate the mean reward function $q$. 
DM then estimates the policy value as
\begin{align*}
    \dm := \mE_{n} [ \mE_{a \sim \pi_e(a|x)} [\hat{q} (x_i, a)] ], 
\end{align*}
where $\hat{q}(x,a)$ is the estimated mean reward function.
If $\hat{q}(x,a)$ is a good approximation to the mean reward function, this estimator accurately estimates the policy value of the evaluation policy.
If $\hat{q}(x,a)$ fails to approximate the mean reward function well, however, the final estimator tends to fail in OPE.

\paragraph{Inverse Probability Weighting (IPW)}
To alleviate the issue with DM, researchers often use IPW~\citep{precup2000eligibility,strehl2010learning}. 
IPW re-weights the rewards by the ratio of the evaluation policy to the behavior policy, as
\begin{align*}
    \ipw := \mE_{n} [ \rho(x_i, a_i) r_i ],
\end{align*}
where $\rho(x, a) := \pi_e(a \mid x) / \pi_b(a \mid x)$ is called the importance weight.
When the behavior policy is known, IPW is unbiased and consistent for the policy value.
However, it can have high variance, especially when the evaluation policy deviates significantly from the behavior policy.
To reduce the variance of IPW, the following weight clipping is often applied.
\begin{align*}
    \ipwps := \mE_{n} [ \min\{\rho(x_i, a_i), \lambda\} r_i ],
\end{align*}
where $\lambda \ge 0$ is a clipping hyperparamter. A lower value of $\lambda$ greatly reduces the variance while introducing a large bias.
Following Su et al.~\citep{su2020doubly}, we call IPW with weight clipping as \textit{IPW with Pessimistic Shrinkage (IPWps)}.
When $\lambda = \infty$, IPWps is identical to IPW.

\paragraph{Doubly Robust (DR)}
DR~\citep{dudik2014doubly} combines DM and IPW as follows.
\begin{align*}
    \dr := \mE_{n} [\mE_{a \sim \pi_e(a|x)} [\hat{q} (x_i, a)] + \rho(x_i, a_i)  (r_i-\hat{q}(x_i, a_i) ) ].
\end{align*}
DR uses the estimated mean reward function as a control variate to decrease the variance of IPW. 
It is also \textit{doubly robust} in that it is consistent to the policy value if either the importance weight or the mean reward estimator is accurate.
The weight clipping can also be applied to DR as follows.
\begin{align*}
    & \drps \\
    & := \mE_{n} [ \mE_{a \sim \pi_e(a|x)} [\hat{q} (x_i, a)] + \min\{\rho(x_i, a_i), \lambda\}  (r_i-\hat{q}(x_i, a_i) ) ],
\end{align*}
where $\lambda \ge 0$ is a clipping hyperparamter. 
DR with weight clipping is called \textit{DR with Pessimistic Shrinkage (DRps)}.
When $\lambda = \infty$, DRps is identical to DR.

\paragraph{Self-Normalized Estimators}
SNIPW~\citep{swaminathan2015self} is an approach to address the variance issue of IPW.
It estimates the policy value by dividing the sum of weighted rewards by the sum of importance weights as:
$$
\snipw :=\frac{\mE_{n} [ \rho(x_i, a_i) r_i ]}{\mE_{n} [ \rho(x_i, a_i) ]}.
$$
SNIPW is more stable than IPW, because the policy value estimated by SNIPW is bounded in the support of rewards, and its conditional variance given action and context is bounded by the conditional variance of the rewards~\citep{kallus2019intrinsically}.
IPW does not have these properties.
We can define Self-Normalized Doubly Robust (SNDR) in a similar manner as follows.
\begin{align*}
    &\sndr \\
    &:=\mE_{n} \left[\mE_{a \sim \pi_e(a|x)} [\hat{q} (x_i, a)] + \frac{\rho(x_i, a_i)}{\mE_{n} [ \rho(x_i, a_i) ]}  (r_i-\hat{q}(x_i, a_i) ) \right].
\end{align*}

\paragraph{Switch Estimator}
DR can still be subject to the variance issue, particularly when the importance weights are large due to low overlap between behavior and evaluation policies.
Switch-DR~\citep{wang2017optimal} aims to further reduce the variance by using DM where the importance weight is large:
\begin{align*} 
    \switchdr
    := \mE_{n} [ & \mE_{a \sim \pi_e(a|x)} [\hat{q} (x_i, a)] \\
    & + \rho(x_i, a_i) \mathbb{I}\{ \rho(x_i, a_i) \le \tau \} (r_i-\hat{q}(x_i, a_i) ) ],
\end{align*}
where $\mathbb{I} \{\cdot\}$ is the indicator function and $\tau \ge 0$ is a hyperparameter.
Switch-DR interpolates between DM and DR. 
When $\tau=0$, it is identical to DM, while $ \tau \to \infty $ yields DR.

\paragraph{Doubly Robust with Optimistic Shrinkage (DRos)}
Su et al.~\citep{su2020doubly} proposes DRos based on a new weight function $\hat{\rho}: \mathcal{X} \times \mathcal{A} \rightarrow \mathbb{R}_{+}$ that directly minimizes sharp bounds on the \textit{mean-squared-error} (MSE) of the resulting estimator. 
DRos is defined as 
\begin{align*}
    & \drs \\
    & := \mE_{n} [\mE_{a \sim \pi_e(a|x)} [\hat{q} (x_i, a)] + \hat{\rho} (x_i, a_i; \lambda)  (r_i-\hat{q}(x_i, a_i) ) ],
\end{align*}
where $\lambda \ge 0$ is a hyperparameter and $\hat{\rho}$ is defined as $\hat{\rho} (x, a; \lambda) := \frac{\lambda}{\rho^{2}(x, a)+\lambda} \rho(x, a)$.
When $\lambda = 0$, $\hat{\rho} (x, a; \lambda) = 0$ leading to DM. 
On the other hand, as $\lambda \rightarrow \infty$, $\hat{\rho} (x, a; \lambda) = \rho(x, a)$ leading to DR.

\paragraph{Cross-Fitting Procedure.}
To obtain a reward estimator, $\hat{q}$, we sometimes use \textit{cross-fitting} to avoid the substantial bias that might arise due to overfitting~\citep{narita2020off}.
The cross-fitting procedure constructs a model-dependent estimator such as DM and DR as follows:
\begin{enumerate}
    \item Take a $K$-fold random partition $\left(\calD_{k}\right)_{k=1}^{K}$ of size $n$ of logged bandit feedback dataset $\calD$ such that the size of each fold is $n_k = n / K$. Also, for each $k=1,2,\ldots K$, we define $\calD_{k}^{c}:=\calD \backslash \calD_{k}$.
    \item For each $k=1,2,\ldots K$, construct reward estimators $\{\hat{q}_k\}_{k=1}^K$ using the subset of data $\calD_{k}^{c}$.
    \item Given $\{\hat{q}_k\}_{k=1}^K$ and model-dependent estimator $\hat{V}$, estimate the policy value by
    $ K^{-1} \sum_{k=1}^K \hat{V} (\pi_e; \calD_{k}, \hat{q}_k) $.
\end{enumerate}

\paragraph{Hyperparameter Tuning Procedure.}
As Table~\ref{tab:hyperparams} summarizes, most OPE estimators have hyperparameters such as $\lambda$, $\tau$, $K$, and $\hat{q}$ that should appropriately be set.
Su et al.~\citep{su2020doubly} proposes to select a set of hyperparameters based on the following criterion.
\begin{align}
    \hat{\theta} \in \argmin_{\theta \in \Theta} \, \mathrm{BiasUB}(\theta; \calD)^2 + \mathbb{V}_n (\theta; \calD),
    \label{eq:hyperparameter_tuning}
\end{align}
where $\mathbb{V}_n (\theta; \calD)$ is the sample variance in OPE, and $\mathrm{BiasUB}(\theta; \calD)$ is the upper bound of the bias estimated using $\calD$. 
There are several ways to derive the bias upper bound as stated in Su et al.~\citep{su2020doubly}.
One way is the direct bias estimation:
\begin{align*}
    \mathrm{BiasUB}(\theta; \calD) 
    & := \left|\mE_n[ \left(\hat{\rho}\left(x_{i}, a_{i}; \theta \right)-\rho\left(x_{i}, a_{i}\right)\right) \left(r_{i}-\hat{q}\left(x_{i}, a_{i}\right)\right) ] \right|  \\
    & +\sqrt{\frac{2 \mathbb{E}\left[\rho (x, a)^{2}\right] \log (2 / \delta)}{n}}+\frac{2 \rho_{\max } \log (2 / \delta)}{3 n}
\end{align*}
where $\delta \in (0, 1]$ is the confidence delta to derive the high probability upper bound, and $\rho_{\max } := \max_{x,a} \rho (x,a)$ is the maximum importance weight.
$\hat{\rho} (x_{i}, a_{i}; \theta )$ is the importance weight modified by a hyperparameter.
For example, for IPWps and DRps, $\hat{\rho} (x_{i}, a_{i}; \lambda ) = \min\{\rho(x_i, a_i), \lambda\}$, 
and for Switch-DR, $\hat{\rho} (x_{i}, a_{i}; \tau ) = \rho(x_i, a_i) \mathbb{I}\{ \rho(x_i, a_i) \le \tau \}$.

\section{Evaluating Offline Evaluation}
So far, we have seen that the OPE community has developed a variety of OPE estimators.
What every OPE research paper should do in their experiments is to compare the performance (estimation accuracy) of the existing estimators and report the results.
A typical and dominant method to do so is to estimate the following \textit{mean-squared-error} (MSE) as the estimator's performance metric:
\begin{align*}
    \MSE := \mE_{\calD} \left[ \left( \trueV - \hat{V} (\pi_e; \calD, \theta) \right)^2  \right],
\end{align*}
where $\trueV$ is the policy value and $\hat{V}$ is an estimator to be evaluated.
MSE measures the squared distance between the policy value and its estimated value; a lower value means a more accurate OPE by $\hat{V}$.
Researchers often calculate the MSE of each estimator several times with different random seeds and report its mean.

The issue with this procedure is that most of the estimators have some hyperparameters that should be chosen properly before the estimation process.
Moreover, the estimation performance can vary when evaluating different evaluation policies (especially in finite sample cases).
However, the current dominant procedure for evaluating OPE estimators uses only one set of hyperparameters and an arbitrary evaluation policy for each estimator, and then discusses the derived results~\citep{wang2017optimal,farajtabar2018more,su2019cab,agarwal2017effective,vlassis2019design}.\footnote{This is why we use $\MSE$ to denote MSE so as to highlight that it depends on the estimator's hyperparameters $\theta$ and an evaluation policy $\pi_e$.}
This type of simplified experimental procedure does not accurately capture the uncertainty in the performance of OPE estimators. 
Specifically, it cannot evaluate the robustness to hyperparameter choices and evaluation policy settings, as the reported score is for a single arbitrary set of hyperparameters and for a single evaluation policy.

What is often critical in offline evaluation practices is to identify an estimator that performs well for a variety of evaluation policies without problem-specific hyperparameter tuning.
An estimator robust to the changes in such configurations is usable reliably in uncertain real-life scenarios.
In contrast, an estimator which performs well only on a narrow set of hyperparameters and evaluation policies entails a higher risk of failure in its particular application.
Therefore, we want to avoid using such sensitive estimators as these estimators are more likely to fail.
In the next section, we describe an experimental procedure that can evaluate the estimators' robustness to experimental configurations, leading to informative estimator comparisons in OPE research and a reliable estimator selection in practice.

\section{Interpretable Evaluation for Offline Evaluation}
Here, we outline our experimental protocol, \textit{Interpretable Evaluation for Offline Evaluation} (IEOE).
As we have discussed, the expected value of performance (e.g., MSE) alone is insufficient to properly evaluate the real-world applicability of an estimator, as it discards information about its robustness to hyperparameter choices and changes in evaluation policies. 
We can conduct a more informative experiment by estimating the \textit{cumulative distribution function} (CDF) of an estimator's performance, as done in some studies on reinforcement learning~\citep{engstrom2020Implementation,jordan2020evaluating,jordan2018using}. 
CDF is the function, $F_{Z}: \mathbb{R} \rightarrow[0,1]$, where $Z$ is a random variable representing the performance metric of an estimator (e.g., the squared error).\footnote{In the following, without loss of generality, we assume that a lower value of $Z$ means more accurate OPE.}
$F_Z(z)$ maps a performance metric $z$ to the probability that the estimator achieves a performance better or equal to that score, i.e., $F_{Z}(z):= \mathbb{P} (Z \leq z)$.

When we have size $m$ of realizations of $Z$, i.e., $\mathcal{Z} := \{z_1, \ldots, z_m\}$, we can estimate the CDF by
\begin{align}
    \hat{F}_{Z}(z):= \frac{1}{m} \sum_{i=1}^m \mathbb{I} \{ z_i \leq z \},
    \label{eq:empirical_cdf}
\end{align}

Using the CDF for evaluating OPE estimators allows researchers to compare different estimators with respect to their robustness to the varying configurations.
Specifically, we can use the CDF to evaluate OPE estimators by examining the CDF of the estimators' performance visually or computing some summary scores of the CDF as the estimators' performance metric.
For example, we can score an estimator by the \textit{area under the CDF curve} (AU-CDF):
$\text{AU-CDF} (z_\mathrm{max}):=\int_{0}^{z_{\mathrm{max}}} F_Z(z) dz.$
Another possible summary score is \textit{conditional value-at-risk} (CVaR) which computes the expected value of a random variable above a given probability $\alpha$:
$\mathrm{CVaR}_{\alpha}(Z):= \mE [Z \mid Z \geq F_{Z}^{-1}(\alpha) ],$ 
where $F_{Z}^{-1}(\alpha):=\argmin_{z} \{z \mid F_{Z}(z) \geq \alpha \}$ is the inverse of the CDF. 
When using CVaR, the estimators are evaluated based on the average performance of the bottom $100 \times (1-\alpha)$ percent of trials. 
For example, $\mathrm{CVaR}_{0.7}(Z)$ is the average performance of the worst 30\% of trials.
In addition, we can use standard deviation (Std), $\mE [ (Z - \mE [Z])^2 ]^{1/2}$, and some other moments such as the skewness of $\hat{F}(z)$ as summary scores.

\begin{figure*}[htb]
\centering
\scalebox{1.}{
\begin{minipage}{0.8\linewidth}
\begin{algorithm}[H]
\caption{Interpretable Evaluation for Offline Evaluation (with Classification Data)}
\begin{algorithmic}[1]
\Require logged bandit feedback $\calD$, an estimator to be evaluated $\hat{V}$, a candidate set of hyperparameters $\Theta$, a set of evaluation policies $\Pi_e$, a hyperparameter sampler $\phi$, a set of random seeds $\calS$
\Ensure empirical CDF of the squared error ($\hat{F}_Z$)
\State $\mathcal{Z} \leftarrow \emptyset$ \Comment{initialize set of results}
\For{$s \in \calS$} 
    \State $ \theta \leftarrow \phi (\Theta; s) $ \Comment{sample a set of hyperparameters}
    \State $ \pi_e \leftarrow \mathrm{Unif} (\Pi_e; s) $ \Comment{sample an evaluation policy uniformly}
    \State $ \calD^* \leftarrow \mathrm{Bootstrap} (\calD; s) $ \Comment{sample logged bandit data \textit{with replacement}}
    \State $z^{\prime} \leftarrow \mathrm{SE} (\hat{V}; \calD^*, \pi_e, \theta) $ \Comment{calculate the squared error of $\hat{V}$}
    \State $ \mathcal{Z} \leftarrow \mathcal{Z} \cup \{z^{\prime}\} $
\EndFor
\State Estimate $F_Z$ using $\mathcal{Z}$ (by Eq.~\ref{eq:empirical_cdf})
\end{algorithmic}
\label{algo:ieoe}
\end{algorithm}
\end{minipage}
}
\end{figure*}

\paragraph{\textbf{IEOE with Synthetic or Classification Data}} 
In research papers, it is common to use synthetic or classification data to evaluate OPE estimators~\citep{dudik2014doubly,wang2017optimal,su2020doubly,kallus2019intrinsically,kallus2020optimal}.
We first present how to apply the IEOE procedure to synthetic or classification data in Algorithm~\ref{algo:ieoe}. 
To evaluate the estimation performance of $\hat{V}$, we need to specify a candidate set of hyperparameters $\Theta$, a set of evaluation policies $\Pi_e$, a hyperparameter sampling function $\phi$, and a set of random seeds $\calS$.
Then, for every seed $s \in \calS$, the algorithm samples a set of hyperparameters $\theta \in \Theta$ based on sampler $\phi$.
What kind of $\phi$ we use can change depending on the purpose of the evaluation of OPE.
For example, we can use a hyperparameter tuning method for OPE estimators such as the method described in Section~\ref{sec:estimators} as $\phi$, assuming practitioners use it in real-world applications.
When we cannot implement such a hyperparameter tuning method for OPE due to its implementation cost or risk of overfitting, we can be conservative and use the uniform distribution as $\phi$ in the evaluation of OPE.
Next, the IEOE algorithm samples an evaluation policy $\pi_e \in \Pi_e$ from the discrete uniform distribution. 
Then, it replicates the data generating process using the \textit{bootstrap sampling} from $\calD$.
A bootstrapped logged bandit feedback dataset is defined as $\calD^* := \{(x_i^*, a_i^*, r_i^*)\}_{i=1}^n$ where each tuple $(x_i^*, a_i^*, r_i^*)$ is sampled independently from $\calD$ \textit{with replacement}. 
Finally, for sampled tuple $(\pi_e, \calD^*, \theta)$, it computes a performance metric (e.g., the squared error).
After applying Algorithm~\ref{algo:ieoe} to several estimators and obtaining the empirical CDF of their evaluation performances, we can visualize them or compute some summary scores to evaluate and compare the estimators' robustness.

\begin{figure*}[!htb]
\centering
\scalebox{0.9}{
\begin{minipage}{0.90\linewidth}
\begin{algorithm}[H]
\caption{Interpretable Evaluation for Offline Evaluation (with Real-World Data)}
\begin{algorithmic}[1]
\Require logged bandit feedback datasets $\{\calD_j\}_{j=1}^\ell$, an estimator to be evaluated $\hat{V}$, a candidate set of hyperparameters $\Theta$, a set of evaluation policies $\Pi_e = \{\pi_j\}_{j=1}^\ell$, a hyperparameter sampler $\phi$, a set of random seeds $\calS$
\Ensure empirical CDF of the squared error ($\hat{F}_Z$)
\State $\mathcal{Z} \leftarrow \emptyset$ (initialize set of results)
\For{$s \in \calS$} 
    \State $ \theta \leftarrow \phi (\Theta; s) $ \Comment{sample a set of hyperparameters based on a given sampler}
    \State $ \pi_j \leftarrow \mathrm{Unif} (\Pi_e; s) $ \Comment{sample an evaluation policy uniformly}
    \State $\calDte = \calD_j$ and $\calDev = \bigcup_{k=1;k\neq j}^\ell \calD_j$ \Comment{define evaluation and test sets}
    \State $ \calDev^* \leftarrow \mathrm{Bootstrap} (\calDev; s) $ \Comment{sample data from the evaluation set \textit{with replacement}}
    \State $ V_{\mathrm{on}} (\pi_j;\calDte) = \mE_n[r_i]$ \Comment{calculate an on-policy estimate of the policy value with the test set}
    \State $z^{\prime} \leftarrow \left(V_{\mathrm{on}} (\pi_j;\calDte) - \hat{V}(\pi_j; \theta, \calDev^*) \right)^2 $ \Comment{calculate the squared error of the estimator}
    \State $ \mathcal{Z} \leftarrow \mathcal{Z} \cup \{z^{\prime}\} $
\EndFor
\State Estimate $F_Z$ using $\mathcal{Z}$ (by Eq.~\ref{eq:empirical_cdf})
\end{algorithmic}
\label{algo:ieoe_real_world}
\end{algorithm}
\end{minipage}
}
\end{figure*}

\paragraph{\textbf{IEOE with Real-World Data}} 
It is also possible to apply IEOE to real-world logged bandit data.
Algorithm~\ref{algo:ieoe_real_world} presents IEOE that can be used in real-world applications. 
To evaluate the performance of $\hat{V}$ with real-world data, we need to prepare several logged bandit feedback datasets $\{\calD_j\}_{j=1}^\ell$ where each dataset $\calD_j$ is collected by a policy $\pi_j$.
Then, for every seed $s \in \calS$, the algorithm samples a set of hyperparameters $\theta \in \Theta$ based on a sampler $\phi$.
Next, the algorithm samples an evaluation policy $\pi_j \in \Pi_e$ from the discrete uniform distribution. 
Then, the evaluation and test sets are defined as $\calDte = \calD_j$ and $\calDev = \bigcup_{k=1;k\neq j}^\ell \calD_j$ where the evaluation set is used in OPE and the test set is used to calculate the ground-truth performance of $\pi_j$.
Then, the algorithm replicates the environment using the \textit{bootstrap sampling} from $\calDev$.
A bootstrapped logged bandit feedback dataset is defined as $\calDev^* := \{(x_i^*, a_i^*, r_i^*)\}_{i=1}^n$ where each tuple $(x_i^*, a_i^*, r_i^*)$ is sampled independently from $\calDev$ \textit{with replacement}. 
Finally, for a sampled tuple $(\pi_e, \calD^*, \theta)$, it computes the squared error as follows.
\begin{align*}
    z= \left(V_{\mathrm{on}} (\pi_j;\calDte) - \hat{V}(\pi_j; \theta, \calDev^*) \right)^2,
\end{align*}
where $V_{\mathrm{on}} (\pi_j;\calDte) = \mE_n[r_i]$ is the on-policy estimate of the policy value of $\pi_j$ estimated with the test set.

Following Algorithm~\ref{algo:ieoe_real_world}, researchers can benchmark the robustness of OPE estimators using public real-world data. 
In addition, practitioners can avoid using unstable estimators by applying Algorithm~\ref{algo:ieoe_real_world} to their own bandit data.

\section{Experiments with Open Bandit Dataset} \label{sec:obd}
In this section, we use IEOE and evaluate the robustness of a wide variety of OPE estimators on Open Bandit Dataset (OBD)\footnote{https://research.zozo.com/data.html}.
We run the experiments using our \textit{pyIEOE} software. 
By using it, anyone can replicate the results easily.\footnote{The code to replicate the results is available at: https://github.com/sony/pyIEOE/benchmark . We also provide detailed description of the software in Appendix~B.}

\subsection{Setup} \label{sec:setup}
OBD is a set of logged bandit feedback datasets collected on a large-scale fashion e-commerce platform provided by Saito et al.~\citep{saito2020open}. 
There are three campaigns, "ALL", "Men", and "Women". 
We use size 30,000 and 300,000 of randomly sub-sampled data from the "ALL" campaign. 
The dataset contains user context as feature vector $x \in \calX$, fashion item recommendation as action $a \in \calA$, and click indicator as reward $r \in \{0,1\}$. 
The dimensions of the feature vector $x$ is 20, and the number of actions is 80.

The dataset consists of subsets of data collected by two different policies, the uniform random policy and the Bernoulli Thompson Sampling policy~\citep{thompson1933likelihood}. 
We let $\calD_{A}$ denote the dataset collected by uniform random policy $\pi_{A}$ and $\calD_{B}$ denote that collected by Bernoulli Thompson Sampling policy $\pi_{B}$. 
We apply Algorithm~\ref{algo:ieoe_real_world} to obtain a set of SEs as the performance metric of the estimators.

\begin{table*}[t]
\begin{tabular}{cc}
\begin{minipage}{1.0\textwidth}
\large
\centering
\caption{Hyperparameter spaces for OPE estimators} 
\label{tab:search_space}
\def\arraystretch{1.2}
\scalebox{0.7}{
\begin{tabular}{c|c}
\toprule
\textbf{OPE Estimators} & \textbf{Hyperparameter Spaces} \\\midrule \midrule
DM & $\hat{q} \in \{\text{LR/RR,RF,LightGBM}\}, K \in \{1,2,\ldots,5\}$ \\
IPWps & $\lambda \in \{1, 5, 10, 50, \ldots, 10^5 \infty\}$, ($\hat{\pi}_b \in \{\text{LR,RF,LightGBM}\}$) \\
SNIPW & ($\hat{\pi}_b \in \{\text{LR,RF,LightGBM}\}$) \\
DRps & $\hat{q} \in \{\text{LR/RR,RF,LightGBM}\}$, $K \in \{1,2,\ldots,5\}$, $\lambda \in \{1, 5, 10, 50, \ldots, 10^5 \infty\}$, ($\hat{\pi}_b \in \{\text{LR,RF,LightGBM}\}$) \\
SNDR & $\hat{q} \in \{\text{LR/RR,RF,LightGBM}\}$, $K \in \{1,2,\ldots,5\}$, ($\hat{\pi}_b \in \{\text{LR,RF,LightGBM}\}$) \\
Switch-DR & $\hat{q} \in \{\text{LR/RR,RF,LightGBM}\}$, $K \in \{1,2,\ldots,5\}$, $\tau \in \{1, 5, 10, 50, \ldots, 10^5 \infty\}$, ($\hat{\pi}_b \in \{\text{LR,RF,LightGBM}\}$) \\
DRos & $\hat{q} \in \{\text{LR/RR,RF,LightGBM}\}$, $K \in \{1,2,\ldots,5\}$,  $\lambda \in \{1, 5, 10, 50, \ldots, 10^5 \infty\}$, ($\hat{\pi}_b \in \{\text{LR,RF,LightGBM}\}$) \\
\bottomrule
\end{tabular}
}
\vskip 0.1in
\raggedright
\fontsize{8.5pt}{8.5pt}\selectfont \textit{Note}: 
LR/RR means that LogisticRegression (LR) is used when $Y$ is binary and RidgeRigression (RR) is used otherwise.
RF stands for RandomForest.
$\hat{\pi}_b$ is an estimated behavior policy.
This is unnecessary when we know the true behavior policy. 
We estimate the behavior policy only in the experiments with classification data in Appendix~A.
Therefore, $\pi_b$ is in parentheses.
$K=1$ means that we do not use cross-fitting and train $\hat{q}$ on the whole $\calDev$.
\end{minipage} 
\\
\\
\begin{minipage}{1.0\textwidth}
\large
\centering
\caption{Hyperparameter spaces for reward estimator $\hat{q}$ (and behavior policy estimator $\hat{\pi}_b$)}
\def\arraystretch{1.2}
\scalebox{0.7}{
\begin{tabular}{c|c}
\toprule
\textbf{Machine Learning Models} & \textbf{Hyperparameter Spaces} \\\midrule \midrule
LogisticRegression (binary outcome) & $C \in [10^{-3}, 10^{3}]$  \\
RidgeRegression (continuous outcome) & $\alpha \in [10^{-2}, 10^{2}]$  \\
RandomForest  & $\text{max\_depth} \in \{2,3,\ldots,10\}$, $\text{min\_samples\_split} \in \{5, 6, \ldots, 20\}$  \\
LightGBM  & $\text{learning\_rate} \in [10^{-4}, 10^{-1}]$, $\text{max\_depth} \in \{2,3,\ldots,10\}$, $\text{min\_samples\_leaf} \in \{5, 6, \ldots, 20\}$  \\
\bottomrule
\end{tabular}
}
\vskip 0.1in
\raggedright
\fontsize{8.5pt}{8.5pt}\selectfont \textit{Note}: 
We follow the \textit{scikit-learn} package as to the names of the hyperparameters
As default, we use $\text{max\_iter}=10,000$ for LogisticRegression, 
$\text{n\_estimators}=100$ for RandomForest,
and $\text{max\_iter}=100$ for LightGBM.
\label{tab:search_space_reward_estimator}
\end{minipage}
\end{tabular}
\end{table*}

\subsection{Estimators and Hyperparameters}
We use our protocol and evaluate DM, IPWps, SNIPW, DRps, SNDR, Switch-DR, and DRos in an interpretable manner.

In the experiment, we use the true behavior policy contained in the dataset to derive importance weights.
In this setting, SNIPW is hyperparameter-free, while the other estimators need to be tested for robustness to the choice of the pre-defined hyperparameters and changes in evaluation policies.
In addition, we use the hyperparameter tuning method described in Section~\ref{sec:estimators} to tune estimator-specific hyperparameters such as $\lambda$ and $\tau$. 
Then, we use RandomizedSearchCV implemented in \textit{scikit-learn} with $\mathrm{n\_iter}=5$ to tune hyperparameters of reward estimator $\hat{q}$.
Tables~\ref{tab:search_space} and~\ref{tab:search_space_reward_estimator} describe hyperparameter spaces $\Theta$ for each estimator.
Finally, we set $\mathcal{S} = \{0,1,\ldots,499\}$.

\begin{figure*}[th]
\begin{center}
    \begin{tabular}{c}
        \begin{minipage}{0.5\hsize}
            \begin{center}
                \includegraphics[clip, width=7.0cm]{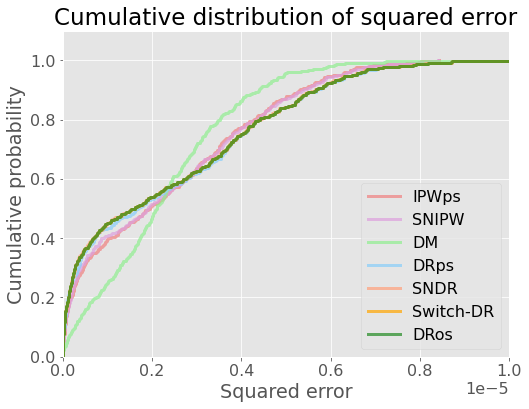} 
                Small ($n=30,000$)
            \end{center}
        \end{minipage}
        
        \begin{minipage}{0.5\hsize}
            \begin{center}
                \includegraphics[clip, width=7.0cm]{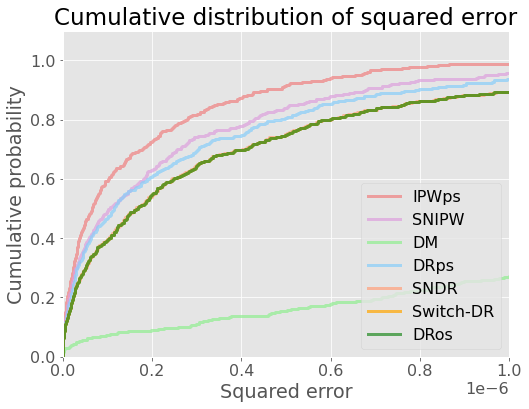}
                Large ($n=300,000$)
            \end{center}
        \end{minipage}
    \end{tabular}
\end{center}
\caption{Comparison of the CDF of OPE estimators' squared error in Open Bandit Dataset}
\label{fig:obd}
\end{figure*}

\begin{table*}[th]
\large
\centering
\caption{Summary scores of the OPE estimators on Open Bandit Dataset with different sample size}
\def\arraystretch{1.2}
\scalebox{0.8}{
\begin{tabular}{c|c|ccccc|cccc}
\toprule
& \multicolumn{4}{c}{$n=30,000$} && \multicolumn{4}{c}{$n=300,000$} \\ \cmidrule{2-5} \cmidrule{7-10}
\textbf{OPE Estimators}  & \textbf{Mean} (typical metric) & \textbf{AU-CDF} & \textbf{CVaR}$_{0.7}$ & \textbf{Std} 
&& \textbf{Mean} (typical metric)  & \textbf{AU-CDF} & \textbf{CVaR}$_{0.7}$ & \textbf{Std} \\
\midrule \midrule
DM 
& \best{1.00} & \best{1.000} & \best{1.00} & \best{1.00}
&& \worst{10.77} & \worst{0.186} & \worst{7.45} & \worst{6.94} \\
IPWps 
& \second{1.02} & \second{0.994} & \second{1.19} & \second{1.31}
&& \best{1.00} & \best{1.000} & \best{1.00} & \best{1.00} \\
SNIPW 
& \second{1.02} & \second{0.994} & 1.20 & 1.33
&& \second{1.58} & \second{0.917} & \second{1.57} & \second{1.71} \\
DRps 
& \worst{1.04} & 0.989 & \worst{1.27} & 1.44
&& 2.48 & 0.887 & 2.67 & 5.02 \\
SNDR 
& \second{1.02} & 0.995 & \worst{1.27} & 1.44
&& 3.27 & 0.827 & 3.50 & 6.01 \\
Switch-DR
& \second{1.02} & \second{0.994} & \worst{1.27} & \worst{1.45}
&& 3.28 & 0.825 & 3.50 & 6.00 \\
DRos 
& \second{1.02} & \second{0.994} & \worst{1.27} & \worst{1.45}
&& 3.28 & 0.825 & 3.50 & 6.00 \\
\bottomrule
\end{tabular}
}
\vskip 0.1in
\raggedright
\fontsize{8.5pt}{8.5pt}\selectfont \textit{Note}:
Larger value is better for \textbf{AU-CDF} and lower value is better for \textbf{Mean}, \textbf{CVaR}, and \textbf{Std}. 
Note that we normalize the scores by dividing them by the best score among all estimators.
We use $z_{\mathrm{max}}=1.0\times10^{-5}$ for $n=30,000$ and $z_{\mathrm{max}}=1.0\times10^{-6}$ for $n=300,000$ to calculate AU-CDF. The \textcolor{dkred}{$\mathbf{red^{\ast}}$} and \textcolor{dkgreen}{$\mathbf{green^{\diamondsuit}}$} fonts represent the best and second-best estimators, respectively. The \textcolor{blue}{$\mathbf{blue^{\dagger}}$} fonts represent the worst estimator.
\label{tab:benchmark_different_size}
\end{table*}

\subsection{Results} \label{app:result}
Figure~\ref{fig:obd} visually compares the CDF of the estimators' squared error. 
Table~\ref{tab:benchmark_different_size} reports AU-CDF, $\text{CVaR}_{0.7}$, and Std as summary scores.

When the dataset size is small ($n = 30,000$), we see that the typical way of reporting only the mean of the squared error cannot tell which estimator is accurate or robust. 
However, some other summary scores show that DM has more robust and stable estimation performance than other estimators, having lower CVaR$_{0.7}$ and Std.
Moreover, Figure~\ref{fig:obd} provides more detailed information about the estimators' performance.
Specifically, DM performs better in the worst case while the other estimators show better performance in the region where squared error is lower than $0.2$.
Thus, when we are conservative and prioritize the worst case performance, DM is the most appropriate choice.
Otherwise, other estimators might be a better choice.
We cannot obtain this conclusion by comparing only the mean (typical metric) of the squared error.

When the dataset size is large ($n = 300,000$), we confirm that IPWps and SNIPW are more accurate than other model-based estimators.
In particular, Figure~\ref{fig:obd} shows that IPWps performs better than other estimators in all region, meaning that we should use it whether we prioritize the best or the worst case performance.

Overall, the results indicate that an appropriate estimator can drastically change depending on the situation such as the data size.
Therefore, we argue that identifying a reasonable estimator before conducting OPE is essential in practice.
Moreover, we demonstrate that the IEOE procedure can provide more informative insight as to the estimators' performance compared to the typical metric.

\section{Real-World Application} \label{sec:real_worl_application}
In this section, we apply the IEOE procedure to a real-world application.

\subsection{Setup}
To show how to use IEOE in a real-world application, we conducted a data collection experiment on a real e-commerce platform in September 2020.
The platform wants to use OPE to improve the performance of its coupon optimization policy safely without conducting A/B tests.
However, it does not know which estimator is appropriate for its specific application and environment.
Therefore, we apply the IEOE procedure with the aim of providing a suitable estimator choice for the platform.

During the data collection experiment, we constructed $\calD_A$, $\calD_B$, and $\calD_C$ by randomly assigning three different policies ($\pi_A$, $\pi_B$, and $\pi_C$) to users on the platform.
In this application, $x$ is a user's context vector, $a$ is a coupon assignment variable (where there are four different types of coupons, i.e., $|\calA|=4$), and $r$ is either a user's content consumption indicator (binary outcome) or the revenue from each user observed within the 7-day period after the coupon assignment (continuous outcome).
The total number of users considered in the experiment was 39,687, and each of $\calD_A$, $\calD_B$, and $\calD_C$ has approximately one third of the users.

Note that, in this application, there is a risk of overfitting due to the intensive hyperparameter tuning of OPE estimators, as the size of the logged bandit feedback data is not large.
Moreover, the data scientists want to use an OPE estimator to evaluate the performance of several candidate policies.
Therefore, we aim to find an estimator that performs stably for a wide range of evaluation policies with fewer hyperparameters.

\subsection{Performance Metric} \label{sec:sony_ieoe}
To apply our evaluation procedure, we need to define a performance metric (in step 8 of Algorithm~\ref{algo:ieoe_real_world}).
We can do this by using our real-world data.
We first pick one of the three policies as evaluation policy $\pi_e$ and regard the others as behavior policies.
When we choose $\pi_A$ as the evaluation policy, we define $\calDev = \calD_B \cup \calD_C$ and $\calDte = \calD_A$.
Then, by applying Algorithm~\ref{algo:ieoe_real_world}, we obtain a set of SEs to evaluate the robustness and real-world applicability of the estimators.

\subsection{Estimators and Hyperparameters}
We use the IEOE protocol to evaluate the robustness of DM, IPWps, SNIPW, DRps, SNDR, Switch-DR, and DRos.
Then, we utilize the experimental results to help the data scientists of the platform choose an appropriate estimator.

During the data collection experiment, we logged the true action choice probabilities of the three policies, and thus SNIPW is hyperparameter-free.
We use the hyperparameter spaces defined in Tables~\ref{tab:search_space} and~\ref{tab:search_space_reward_estimator} for our real-world application.
In addition, we use the hyperparameter tuning method described in Section~\ref{sec:estimators} to tune estimator-specific hyperparameters such as $\lambda$ and $\tau$. 
Then, we use the uniform distribution as $\phi$ to sample hyperparameters of reward regression model $\hat{q}$.
Finally, we set $\mathcal{S} = \{0,1,\ldots,999\}$ and $\Pi_e =\{\pi_A, \pi_B, \pi_C\}$.

\begin{figure*}[t]
\begin{center}
    \begin{tabular}{c}
        \begin{minipage}{0.5\hsize}
            \begin{center}
                \includegraphics[clip, width=7.0cm]{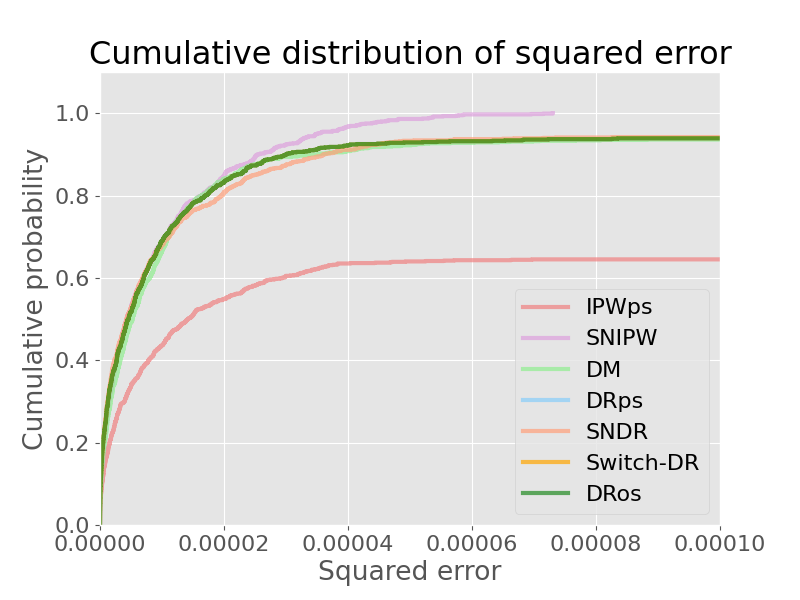} 
                binary
            \end{center}
        \end{minipage}
        
        \begin{minipage}{0.5\hsize}
            \begin{center}
                \includegraphics[clip, width=7.0cm]{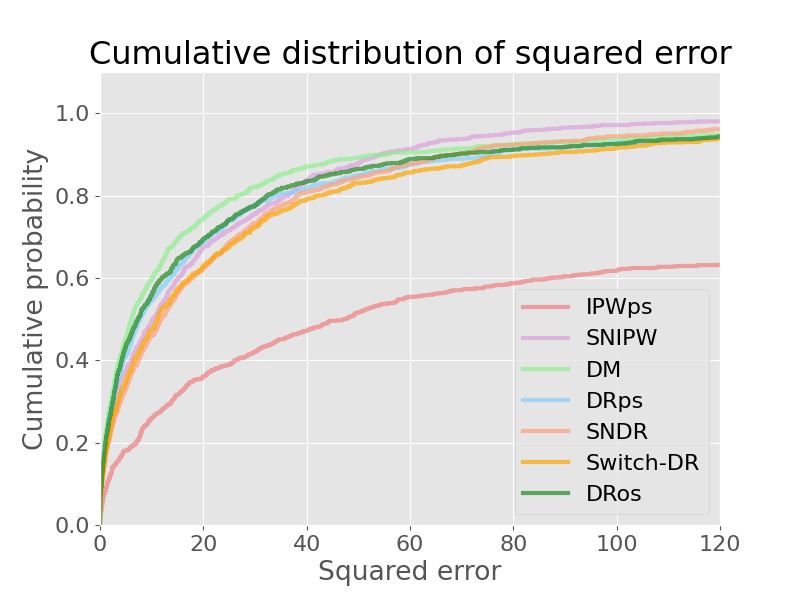}
                continuous
            \end{center}
        \end{minipage}
    \end{tabular}
\end{center}
\caption{Comparison of the CDF of OPE estimators' squared error in the real-world application}
\label{fig:sony_cdf}
\end{figure*}

\begin{table*}[t]
\large
\centering
\caption{Summary scores of the OPE estimators in the real-world application}
\label{tab:sony}
\def\arraystretch{1.2}
\scalebox{0.75}{
\begin{tabular}{c|c|ccccc|cccc}
\toprule
& \multicolumn{4}{c}{Binary Outcome} && \multicolumn{4}{c}{Continuous Outcome} \\ \cmidrule{2-5} \cmidrule{7-10}
\textbf{OPE Estimators}  & \textbf{Mean} (typical metric) & \textbf{AU-CDF} & \textbf{CVaR}$_{0.7}$ & \textbf{Std} 
&& \textbf{Mean} (typical metric)  & \textbf{AU-CDF} & \textbf{CVaR}$_{0.7}$ & \textbf{Std} \\
\midrule \midrule
DM 
& 8.70 & 0.946 & 10.92 & \worst{35.94}
&& 1.29 & \best{1.000} & 1.47 & 2.19 \\
IPWps 
& \worst{29.45} & \worst{0.648} & \worst{31.96} & \second{29.84}
&& \worst{19.00} & \worst{0.572} & \worst{19.84} & \worst{14.67} \\
SNIPW 
& \best{1.00} & \best{1.000} & \best{1.00} & \best{1.00}
&& \best{1.00} & \second{0.974} & \best{1.00} & \best{1.00} \\
DRps 
& 8.16 & \second{0.953} & 10.27 & 34.54
&& 1.44 & 0.957 & 1.60 & 2.11 \\
SNDR 
& \second{7.45} & 0.942 & \second{9.35} & 32.19
&& 1.21 & 0.935 & \second{1.22} & \second{1.17} \\
Switch-DR 
& 8.16 & \second{0.953} & 10.27 & 34.54
&& 1.48 & 0.919 & 1.57 & 1.68 \\
DRos 
& 8.16 & \second{0.953} & 10.27 & 34.54
&& 1.43 & 0.968 & 1.60 & 2.21 \\
\bottomrule
\end{tabular}
}
\vskip 0.1in
\raggedright
\fontsize{8.5pt}{8.5pt}\selectfont \textit{Note}:
\textbf{Binary Outcome} is the results when the outcome is each user's content consumption indicator. 
\textbf{Continuous Outcome} is the results when the outcome is the revenue from each user observed within the 7-day period after the coupon assignment.
Larger value is better for \textbf{AU-CDF} and lower value is better for \textbf{Mean}, \textbf{CVaR}, and \textbf{Std}.
Note that we normalize the scores by dividing them by the best score among all estimators.
We use $z_{\mathrm{max}}=1.0\times10^{-4}$ for the binary outcome and $z_{\mathrm{max}}=1.0\times10^2$ for the continuous outcome to calculate AU-CDF. The \textcolor{dkred}{$\mathbf{red^{\ast}}$} and \textcolor{dkgreen}{$\mathbf{green^{\diamondsuit}}$} fonts represent the best and second-best estimators, respectively. The \textcolor{blue}{$\mathbf{blue^{\dagger}}$} fonts represent the worst estimator.
\end{table*}

\subsection{Results}
We applied Algorithm~\ref{algo:ieoe_real_world} to the above estimators for the binary and continuous outcome data, respectively.

Figure~\ref{fig:sony_cdf} compares the CDF of the estimators' squared error for each outcome.
First, it is obvious that SNIPW is the best estimator for the binary outcome case, achieving the best accuracy in almost all regions.
We can also argue that SNIPW is preferable for the continuous outcome case, because it reveals the most accurate estimation in the worst case and is hyperparameter-free, although it underperforms DM in some cases.
On the other hand, IPWps performs poorly for both outcomes, because our dataset is not large and some behavior policies are near deterministic, making IPWps an unstable estimator.
Moreover, Switch-DR fails to accurately evaluate the performance of the evaluation policies. 
Thus, it is unsafe to use these estimators in our application, even though we tune their hyperparameters ($\lambda$ or $\tau$).

We additionally confirm the above observations in a quantitative manner.
For both binary and continuous outcomes, we compute AU-CDF, CVaR$_{0.7}$, and Std of the squared error for each OPE estimator.
We report these summary scores in Table~\ref{tab:sony}, and the results demonstrate that SNIPW clearly outperforms other estimators in almost all situations.
In particular, SNIPW is the best with respect to CVaR$_{0.7}$ and Std for both binary and continuous outcomes, showing that this estimator is the most stable estimator in our environment.
Moreover, SNIPW is hyperparameter-free, and overfitting is less likely to occur compared to other estimators having some hyperparameters to be tuned.
Through this evaluation of OPE estimators, we concluded that \textbf{\textit{the e-commerce platform should use SNIPW for its offline evaluation.}}
After comprehensive accuracy and stability verification, the platform is now using SNIPW to improve its coupon optimization policy safely.

\section{Conclusion and Future Work}
In this paper, we argued that the current dominant evaluation procedure for OPE cannot evaluate the robustness of the estimators' performance.
Instead, the IEOE procedure can provide an interpretable way to evaluate how robust each estimator is to the choice of hyperparameters or changes in evaluation policies.
We have also developed open-source software to streamline our interpretable evaluation procedure.
It enables rapid benchmarking and validation of OPE estimators so that practitioners can spend more time on real decision making problems, and OPE researchers can focus more on tackling advanced technical questions.
We perform an extensive evaluation of a wide variety of OPE estimators and demonstrated that our experiments are more informative than a typical procedure, showing which estimators are more sensitive to configuration changes.
Finally, we applied our procedure to a real-world application and demonstrated its practical usage.

Although our procedure is useful to evaluate the robustness of estimators, we need to prepare at least two logged bandit feedback datasets collected by different policies to apply it to real-world applications, as described in Algorithm~\ref{algo:ieoe_real_world}.
Thus, it would be beneficial to construct a procedure to enable the evaluation of OPE estimators with only logged bandit data collected by a single policy.

\begin{acks}
The authors would like to thank Masahiro Nomura, Ryo Kuroiwa, and Richard Liu for their help in reviewing the paper.
Additionally, we would like to thank the anonymous reviewers for their constructive reviews and discussions.
\end{acks}

\bibliographystyle{ACM-Reference-Format}
\bibliography{main.bbl}

\clearpage
\input{appendix}

\end{document}

%% file: notation.tex
\newcommand{\mE}{\mathbb{E}}

\newcommand{\calD}{\mathcal{D}}
\newcommand{\calX}{\mathcal{X}}
\newcommand{\calA}{\mathcal{A}}

\newcommand{\calS}{\mathcal{S}}

\newcommand{\trueV}{V(\pi_e)}
\newcommand{\dm}{\hat{V}_{\mathrm{DM}} (\pi_e; \calD, \hat{q})}
\newcommand{\ipw}{\hat{V}_{\mathrm{IPW}} (\pi_e; \calD)}
\newcommand{\ipwps}{\hat{V}_{\mathrm{IPWps}} (\pi_e; \calD)}
\newcommand{\snipw}{\hat{V}_{\mathrm{SNIPW}} (\pi_e; \calD)}

\newcommand{\dr}{\hat{V}_{\mathrm{DR}} (\pi_e; \calD, \hat{q})}
\newcommand{\drps}{\hat{V}_{\mathrm{DRps}} (\pi_e; \calD, \hat{q})}
\newcommand{\sndr}{\hat{V}_{\mathrm{SNDR}} (\pi_e; \calD, \hat{q})}
\newcommand{\drs}{\hat{V}_{\mathrm{DRos}} (\pi_e; \calD, \hat{q}, \lambda)}

\newcommand{\switchdr}{\hat{V}_{\mathrm{SwitchDR}} (\pi_e; \calD, \hat{q}, \tau)}

\newcommand{\calDev}{\calD_{\mathrm{ev}}}
\newcommand{\calDte}{\calD_{\mathrm{te}}}

\newcommand{\MSE}{\mathrm{MSE} (\hat{V}; \pi_e, \theta)}

\newcommand{\argmin}{\mathop{\rm argmin}\limits}

\usepackage{listings}
\definecolor{dkred}{rgb}{0.8,0,0}
\definecolor{dkgreen}{rgb}{0,0.4,0}
\definecolor{gray}{rgb}{0.2,0.2,0.2}
\definecolor{mauve}{rgb}{0.7,0,0.9}
\newcommand{\best}[1]{\textcolor{dkred}{$\mathbf{{#1}^{\ast}}$}}
\newcommand{\second}[1]{\textcolor{dkgreen}{$\mathbf{{#1}^{\diamondsuit}}$}}
\newcommand{\worst}[1]{\textcolor{blue}{$\mathbf{{#1}^{\dagger}}$}}

%% file: appendix.tex
\appendix
\onecolumn

\begin{table*}[t]
\large
\centering
\caption{Classification datasets used in the benchmark experiment}
\def\arraystretch{1.2}
\scalebox{0.8}{
\begin{tabular}{c|ccccc}
\toprule
\textbf{Datasets} && \textbf{\#Samples} & \textbf{\#Actions} & \textbf{\#Dimensions} \\\midrule \midrule
OptDigits && 5,620 & 10 & 64 \\
PenDigits && 10,992 & 10 & 16 \\
SatImage && 6,435 & 6 & 36 \\
\bottomrule
\end{tabular}
}
\vskip 0.1in
\raggedright
\fontsize{8.5pt}{8.5pt}\selectfont \textit{Note}: 
\textbf{\#Samples} is the size of the dataset. 
\textbf{\#Actions} is the total number of actions (or classes).
\textbf{\#Dimensions} is the number of dimensions of the context (or feature) vector.
\label{tab:data}
\end{table*}

\begin{table*}[t]
\large
\centering
\caption{Behavior and evaluation policies used in the benchmark experiment}
\label{tab:base_policies}
\def\arraystretch{1.2}
\scalebox{0.8}{
\begin{tabular}{c|cc}
\toprule
\textbf{Behavior and Evaluation Policies} & \textbf{Base Machine Learning Classifier} ($\pi_{det}$) & \textbf{Alpha} ($\alpha$) \\\midrule \midrule
behavior policy & LogisticRegression & 0.9 \\
evaluation policy 1 & LogisticRegression & 0.8 \\
evaluation policy 2 & LogisticRegression & 0.2 \\
evaluation policy 3 & RandomForest & 0.8 \\
evaluation policy 4 & RandomForest & 0.2 \\
evaluation policy 5 & None (uniform random) & 0.0 \\
\bottomrule
\end{tabular}
}
\vskip 0.1in
\raggedright
\fontsize{8.5pt}{8.5pt}\selectfont \textit{Note}: 
For LogisticRegression, we use $\text{C}=100, \text{max\_iter}=10000$. For RandomForest, we use $\text{n\_estimators}=100$, $\text{min\_samples\_split}=5, \text{max\_depth}=10$. We also set $\text{random\_state}=12345$ for both classifiers. The names of the hyperparameters correspond to the ones specified by the \textit{scikit-learn} package.
\end{table*}

\begin{figure*}[h]
\begin{tabular}{cccc}
\toprule
& \multicolumn{3}{c}{Datasets} \\ \midrule
& Optdigits & Pendigits & Satimage  \\ \midrule \midrule
zoom &
\begin{minipage}{0.30\hsize}
    \begin{center}
        \includegraphics[clip, width=4.5cm]{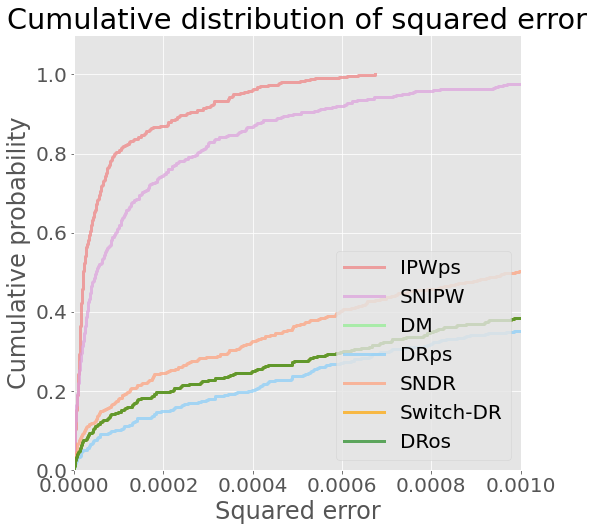}
    \end{center}
\end{minipage}
&
\begin{minipage}{0.30\hsize}
    \begin{center}
        \includegraphics[clip, width=4.5cm]{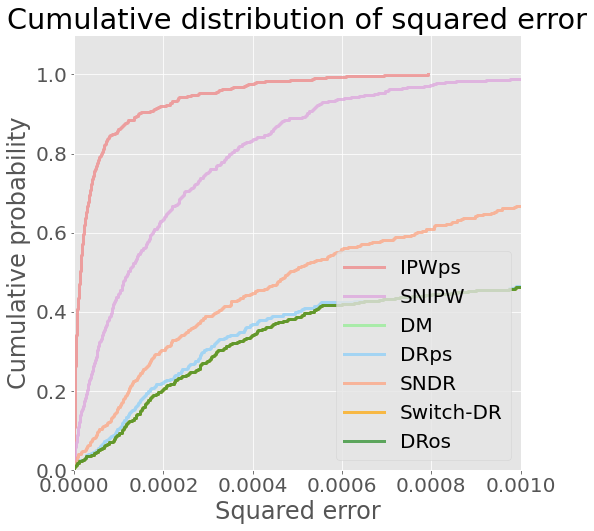}
    \end{center}
\end{minipage}
&
\begin{minipage}{0.30\hsize}
    \begin{center}
        \includegraphics[clip, width=4.5cm]{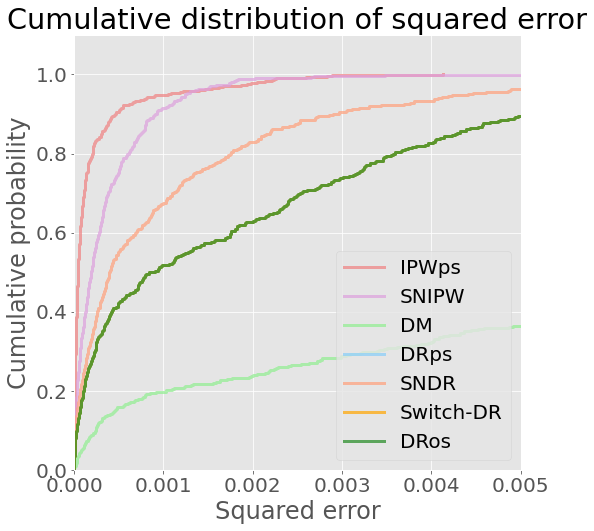}
    \end{center}
\end{minipage}
\\
full &
\begin{minipage}{0.30\hsize}
    \begin{center}
        \includegraphics[clip, width=4.5cm]{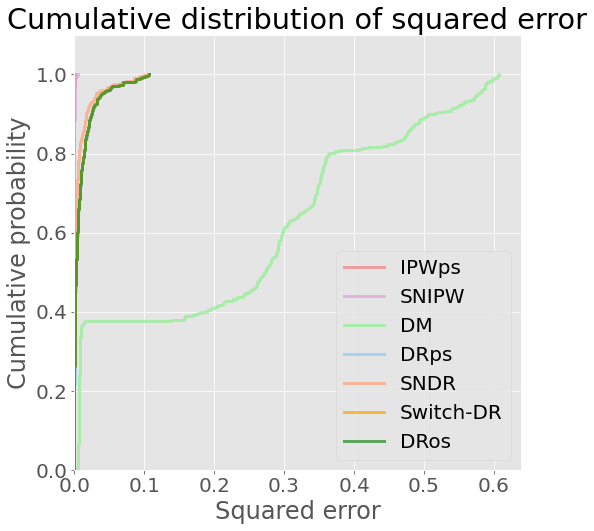}
    \end{center}
\end{minipage}
&
\begin{minipage}{0.30\hsize}
    \begin{center}
        \includegraphics[clip, width=4.5cm]{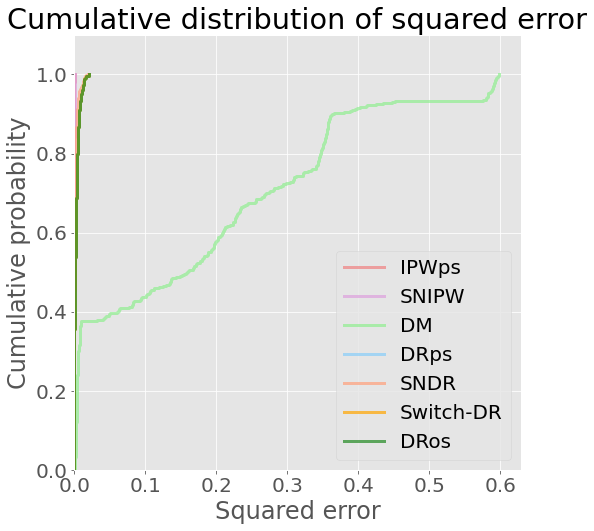}
    \end{center}
\end{minipage}
&
\begin{minipage}{0.30\hsize}
    \begin{center}
        \includegraphics[clip, width=4.5cm]{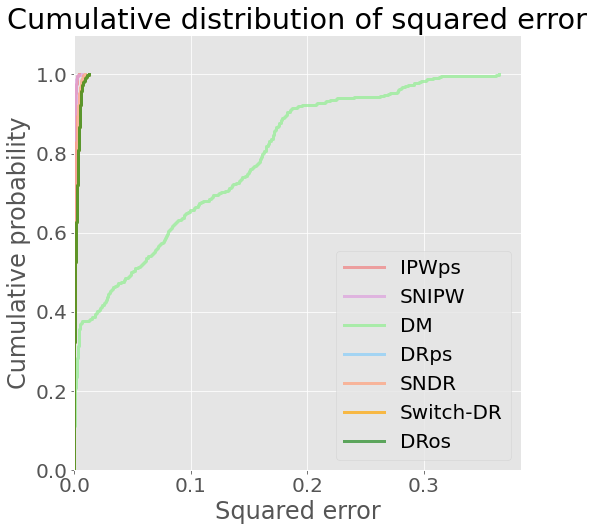}
    \end{center}
\end{minipage}
\\
\multicolumn{4}{c}{
\begin{minipage}{1.0\hsize}
\begin{center}
\caption{Comparison of the CDF of OPE estimators’ squared error (\textbf{true behavior policy})}
\label{fig:multiclass_true}
\end{center}
\end{minipage}
}
\\ \midrule
zoom &
\begin{minipage}{0.30\hsize}
    \begin{center}
        \includegraphics[clip, width=4.5cm]{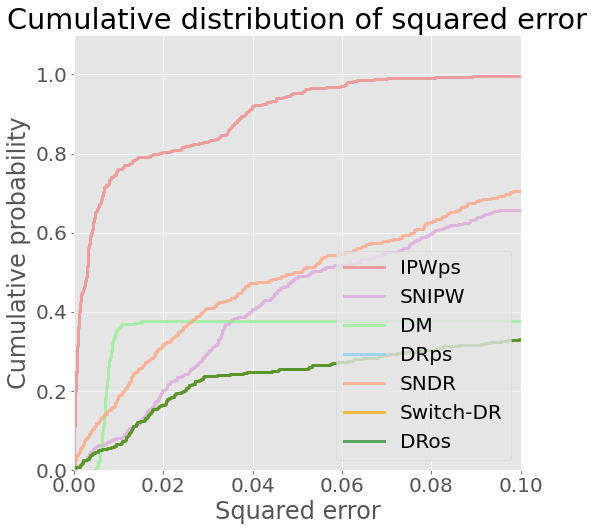}
    \end{center}
\end{minipage}
&
\begin{minipage}{0.30\hsize}
    \begin{center}
        \includegraphics[clip, width=4.5cm]{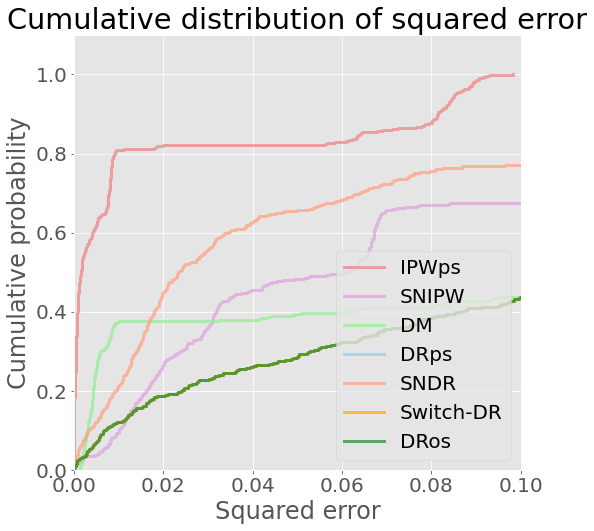}
    \end{center}
\end{minipage}
&
\begin{minipage}{0.30\hsize}
    \begin{center}
        \includegraphics[clip, width=4.5cm]{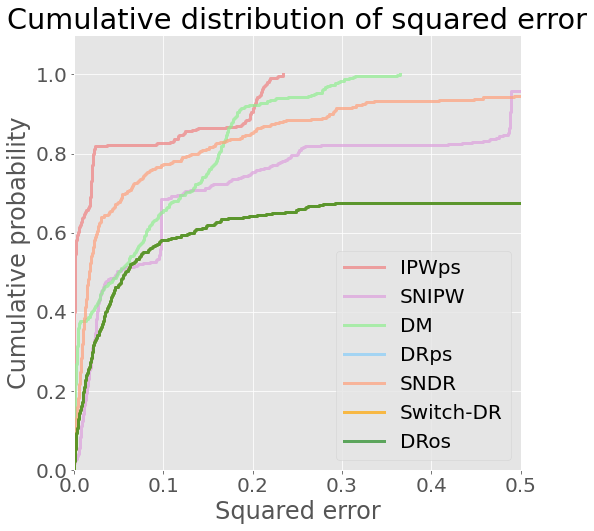}
    \end{center}
\end{minipage}
\\
full &
\begin{minipage}{0.30\hsize}
    \begin{center}
        \includegraphics[clip, width=4.5cm]{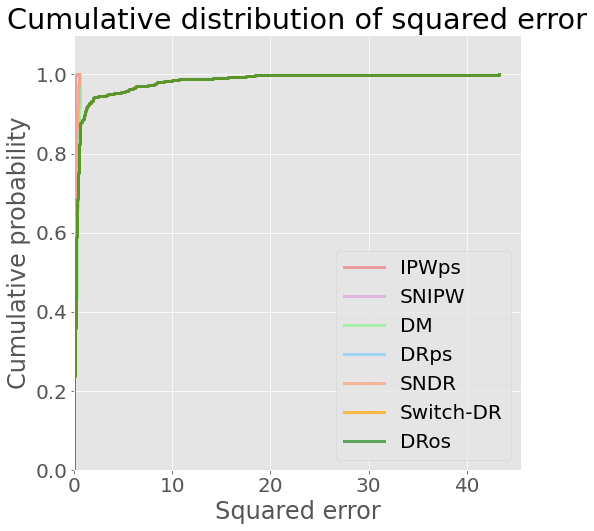}
    \end{center}
\end{minipage}
&
\begin{minipage}{0.30\hsize}
    \begin{center}
        \includegraphics[clip, width=4.5cm]{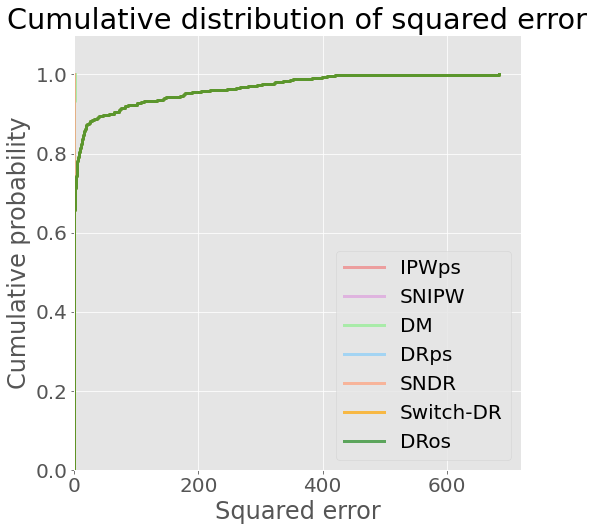}
    \end{center}
\end{minipage}
&
\begin{minipage}{0.30\hsize}
    \begin{center}
        \includegraphics[clip, width=4.5cm]{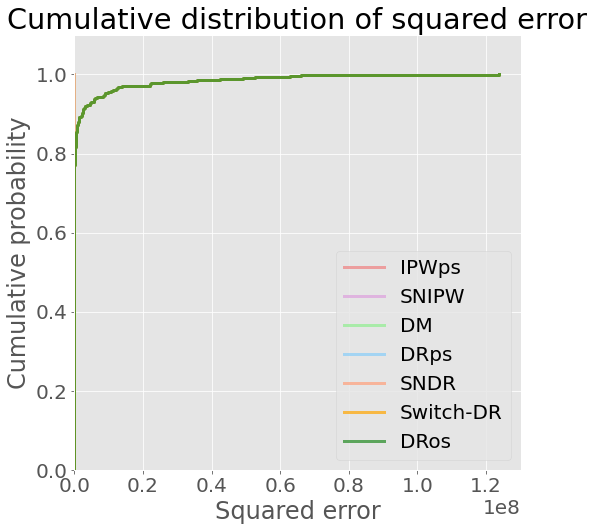}
    \end{center}
\end{minipage}
\\
\multicolumn{4}{c}{
\begin{minipage}{1.0\hsize}
\begin{center}
\caption{Comparison of the CDF of OPE estimators’ squared error (\textbf{estimated behavior policy})}
\label{fig:multiclass_estimated}
\end{center}
\end{minipage}
}
\\ \bottomrule
\end{tabular}
\end{figure*}

\begin{table*}[t]
\begin{tabular}{cc}
\begin{minipage}{1.0\textwidth}
\large
\centering
\caption{Summary scores of the OPE estimators (\textbf{true behavior policy})}
\def\arraystretch{1.2}
\scalebox{0.75}{
\begin{tabular}{c|cccccccccccc}
\toprule
 & \multicolumn{3}{c}{OptDigits} && \multicolumn{3}{c}{PenDigits} && \multicolumn{3}{c}{SatImage} \\ \cmidrule{2-4} \cmidrule{6-8} \cmidrule{10-12}
\textbf{OPE Estimators} & \textbf{AU-CDF} & \textbf{CVaR}$_{0.7}$ & \textbf{Std} && \textbf{AU-CDF} & \textbf{CVaR}$_{0.7}$ & \textbf{Std} && \textbf{AU-CDF} & \textbf{CVaR}$_{0.7}$ & \textbf{Std} \\\midrule \midrule
DM 
& \worst{0.000} & \worst{2215.37} & \worst{1631.15}
&& \worst{0.000} & \worst{2609.92} & \worst{1674.67}
&& \worst{0.266} & \worst{323.81} & \worst{184.39} \\
IPWps 
& \best{1.000} & \best{1.00} & \best{1.00}
&& \best{1.000} & \best{1.00} & \best{1.00}
&& \best{1.000} & \best{1.00} & \best{1.00} \\
SNIPW 
& \second{0.907} & \second{2.51} & \second{3.15}
&& \second{0.843} & \second{3.16} & \second{2.18}
&& \second{0.966} & \second{1.56} & \second{1.16} \\
DRps 
& 0.249 & 127.61 & 139.99
&& 0.358 & 45.49 & 33.59
&& 0.686 & 7.99 & 4.78 \\
SNDR 
& 0.374 & 96.91 & 125.92 
&& 0.480 & 27.47 & 26.15
&& 0.833 & 4.71 & 3.17 \\
Switch-DR 
& 0.287 & 126.21 & 139.12
&& 0.347 & 45.46 & 33.52
&& 0.686 & 7.99 & 4.78 \\
DRos 
& 0.287 & 126.21 & 139.12
&& 0.347 & 45.46 & 33.52
&& 0.686 & 7.99 & 4.78 \\
\bottomrule
\end{tabular}
}
\vskip 0.1in
\raggedright
\fontsize{8.5pt}{8.5pt}\selectfont \textit{Note}:
Larger value is better for \textbf{AU-CDF} and lower value is better for \textbf{CVaR} and \textbf{Std}. 
Note that we normalize the summary scores by dividing them by the best score among all estimators.
We use $z_{\mathrm{max}}=1.0\times10^{-3}$ for OptDigits and Pendigits and $z_{\mathrm{max}}=5.0\times10^{-3}$ for SatImage to calculate AU-CDF. The \textcolor{dkred}{$\mathbf{red^{\ast}}$} and \textcolor{dkgreen}{$\mathbf{green^{\diamondsuit}}$} fonts represent the best and second-best estimators, respectively. The \textcolor{blue}{$\mathbf{blue^{\dagger}}$} fonts represent the worst estimator.
\label{tab:multiclass_true}
\end{minipage}
\\
\\
\begin{minipage}{1.0\textwidth}
\large
\centering
\caption{Summary scores of the OPE estimators (\textbf{estimated behavior policy})}
\def\arraystretch{1.2}
\scalebox{0.75}{
\begin{tabular}{c|cccccccccccc}
\toprule
 & \multicolumn{3}{c}{OptDigits} && \multicolumn{3}{c}{PenDigits} && \multicolumn{3}{c}{SatImage} \\ \cmidrule{2-4} \cmidrule{6-8} \cmidrule{10-12}
\textbf{OPE Estimators} & \textbf{AU-CDF} & \textbf{CVaR}$_{0.7}$ & \textbf{Std} && \textbf{AU-CDF} & \textbf{CVaR}$_{0.7}$ & \textbf{Std} && \textbf{AU-CDF} & \textbf{CVaR}$_{0.7}$ & \textbf{Std} \\\midrule \midrule
DM 
& 0.390 & 14.18 & 9.89
&& 0.454 & 7.84 & 5.86
&& \second{0.911} & \second{1.62} & \second{1.20} \\
IPWps 
& \best{1.000} & \best{1.00} & \best{1.00}
&& \best{1.000} & \best{1.00} & \best{1.00}
&& \best{1.000} & \best{1.00} & \best{1.00} \\
SNIPW 
& 0.460 & \second{9.06} & 7.05
&& 0.538 & 7.22 & 5.23
&& 0.778 & 3.23 & 2.52 \\
DRps 
& \worst{0.262} & \worst{71.18} & \worst{141.25}
&& \worst{0.332} & \worst{1527.07} & \worst{2498.85}
&& \worst{0.657} & \worst{5.27 \times 10^7} & \worst{1.25 \times 10^8} \\
SNDR 
& \second{0.525} & 7.57 & \second{6.07}
&& \second{0.700} & \second{4.28} & \second{3.79}
&& 0.904 & 2.40 & 2.52 \\
Switch-DR 
& \worst{0.262} & \worst{71.18} & \worst{141.25}
&& \worst{0.332} & \worst{1527.07} & \worst{2498.85}
&& \worst{0.657} & \worst{5.27 \times 10^7} & \worst{1.25 \times 10^8} \\
DRos 
& \worst{0.262} & \worst{71.18} & \worst{141.25}
&& \worst{0.332} & \worst{1527.07} & \worst{2498.85}
&& \worst{0.657} & \worst{5.27 \times 10^7} & \worst{1.25 \times 10^8} \\
\bottomrule
\end{tabular}
}
\vskip 0.1in
\raggedright
\fontsize{8.5pt}{8.5pt}\selectfont \textit{Note}:
Larger value is better for \textbf{AU-CDF} and lower value is better for \textbf{CVaR} and \textbf{Std}.
Note that we normalize the scores by dividing them by the best score among all estimators.
We use $z_{\mathrm{max}}=0.1$ for OptDigits and Pendigits and $z_{\mathrm{max}}=0.5$ for SatImage to calculate AU-CDF. The \textcolor{dkred}{$\mathbf{red^{\ast}}$} and \textcolor{dkgreen}{$\mathbf{green^{\diamondsuit}}$} fonts represent the best and second-best estimators, respectively. The \textcolor{blue}{$\mathbf{blue^{\dagger}}$} fonts represent the worst estimator.
\label{tab:multiclass_estimated}
\end{minipage}
\end{tabular}
\end{table*}

\section{Benchmark Experiments on Classification Datasets} \label{app:benchmark}
Here, we conduct experiments on three classification datasets, OptDigits, PenDigits, and SatImage provided at the UCI repository~\citep{dua2017uci}. 
Table~\ref{tab:data} shows some statistics of the datasets used in the benchmark experiment.

\subsection{Setup}
Following previous studies~\citep{farajtabar2018more,dudik2014doubly,wang2017optimal,kallus2020optimal}, we transform classification data to contextual bandit feedback data. In a classification dataset $\{(x_i, a_i)\}_{i=1}^{n}$, we have feature vector $x_i \in \calX$ and ground-truth label $a_i \in \calA$. Here, we regard a machine learning classifier $\pi_{det}: \calX \rightarrow \Delta(\calA)$ as a deterministic policy that chooses class label $a_i \in \calA$ as an action from feature vector $x_i$. We then define reward variable $r_i := \mathbb{I}\{\pi(x_i) = a_i \}$. Since the original classifier is deterministic, we make it stochastic by combining $\pi_{det}$ and the uniform random policy $\pi_u$ as:
$$\pi(a \mid x) = \alpha \cdot \mathbb{I} \{ \pi_{det}(x) =a \} + (1 - \alpha) \cdot \pi_u(a),$$ 
where $\alpha \in [0,1]$ is an additional experimental setting.

To apply IEOE to classification data, we first randomly split each dataset into train $\calD_{\text{tr}}$ and test $\calDte := \{(x_i, a_i)\}_{i=1}^{n^{\prime}}$ sets. Then, we train a classifier on $\calD_{\text{tr}}$, and use it to construct a behavior policy $\pi_b$ and a class of evaluation policies $\Pi_e$. By running behavior policy $\pi_b$ on $\calDte$, we transform $\calDte$ to logged bandit feedback data $\calDev := \{(x_i, a_i^b, r_i=\mathbb{I}\{a_i^b = a_i \})\}_{i=1}^{n^{\prime}}$, where $a_i^b \sim \pi_b$ is the action sampled by the behavior policy. Then, by applying the following procedure, we compute the squared error (SE) of $\hat{V}$ for each iteration in Algorithm~\ref{algo:ieoe}:
\begin{enumerate}
    \item Estimate the policy value $\hat{V}(\pi_e; \calD^*, \theta)$ for tuple $(\pi_e, \calD^*, \theta)$ sampled in the algorithm.
    \item Estimate $V(\pi_e)$ using the fully observed rewards in $\calDte$, i.e., $V(\pi_e ; \calDte) := \mE_{n^{\prime}} [ \mE_{a^e \sim \pi_e(a|x_i)} [\mathbb{I}\{a^e = a_i\}]]$.
    \item Compare the off-policy estimate $\hat{V} (\pi_e; \calD^*, \theta)$ with its ground-truth $V(\pi_e ; \calDte)$ using SE as a performance metric of $\hat{V}$, i.e., $\mathrm{SE} (\hat{V}; \calD^*, \pi_e, \theta) := (\hat{V} (\pi_e; \calD^*, \theta) -  V(\pi_e ; \calDte) )^2$.
\end{enumerate}

\subsection{Estimators and Hyperparameters}
We use IEOE to evaluate the robustness of DM, IPWps, SNIPW, DRps, SNDR, Switch-DR, and DRos.

Here, we run the experiments under two different settings. 
First, we test the case where the true behavior policy $\pi_b$ is available.
Next, we investigate the OPE estimators with estimated behavior policy $\hat{\pi}_b$, where we assume that the true behavior policy is unknown. 
In this case, we additionally test the OPE estimators for robustness to the choice of machine learning method to obtain $\hat{\pi}_b$.

Tables~\ref{tab:search_space} and~\ref{tab:search_space_reward_estimator} (in the main text) describe hyperparameter spaces $\Theta$ for each estimator.
We use RandomizedSearchCV implemented in \textit{scikit-learn} with $\mathrm{n\_iter}=5$ to tune hyperparameters of reward estimator $\hat{q}$ and behavior policy estimator $\hat{\pi}_b$. 
We additionally use CalibratedClassifierCV implemented in \textit{scikit-learn} with $\mathrm{cv}=2$ when estimating the behavior policy, as calibrating the behavior policy estimator matters in OPE~\citep{raghu2018behaviour}.
Then, we use the hyperparameter tuning method described in Section~\ref{sec:estimators} to tune estimator-specific hyperparameters such as $\lambda$ and $\tau$. 
Table~\ref{tab:base_policies} describes how we construct the true behavior policy and five different evaluation policies in $\Pi_e$.
Finally, we set $\mathcal{S} = \{0,1,\ldots,499\}$.

\subsection{Results}

Figures~\ref{fig:multiclass_true} and~\ref{fig:multiclass_estimated} visually compare the CDF of the estimators' squared error for each dataset in true and estimated behavior policy settings. We also confirm the observations in a quantitative manner by computing AU-CDF, $\text{CVaR}_{0.7}$, and Std of the squared error of each OPE estimator. We report these summary scores in Tables~\ref{tab:multiclass_true} and~\ref{tab:multiclass_estimated}.

First, in the setting where the true behavior policy is available, it is obvious that IPWps is the best estimator and achieves the most accurate estimation in almost all regions (see Figure~\ref{fig:multiclass_true}). 
SNIPW also performs comparably better than other estimators. 
In contrast, model-dependent estimators, especially DM, perform poorly compared to the typical estimators such as IPWps and SNIPW. 
We observe here that these model-dependent estimators perform worse, when the reward estimator $\hat{q}$ has a serious bias issue.
On the other hand, we do not have to care about the specification of $\hat{q}$ when we use IPWps or SNIPW.
Therefore, our experimental procedure poses a possibility that simple estimators with fewer hyperparameters tend to perform well and be robust for a wide variety of settings when the true behavior policy is recorded.

In the setting where the behavior policy needs to be estimated, we observe similar trends.
First, Figure~\ref{fig:multiclass_estimated} and Table~\ref{tab:multiclass_estimated} show that IPWps achieves the most accurate estimation even when it uses the estimated behavior policy.
Second, estimators based on DR such as DRps, Switch-DR, and DRos show considerably large squared errors when the behavior policy is estimated.
This is because DR is vulnerable to the overfitting of $\hat{\pi}_b$.
DR produces large squared errors when $\hat{\pi}_b$ overfits the data and outputs extreme estimations (we observe that the minimum estimated action choice probability is $10^{-7}$). 
With these extreme estimated action choice probabilities, the importance weights used in these estimators also become large, amplifying the estimation error of reward estimator $\hat{q}$.
This leads to serious overestimation of the policy value of $\pi_e$, even though the cut-off hyperparameters ($\lambda$ and $\tau$) are properly tuned.

We suggest that future OPE research use the IEOE procedure to test the stability and robustness of OPE estimators as we have demonstrated. 
This additional experimental effort will produce substantial information about the estimators' usability in practice.

\clearpage
\section{Software Implementation} \label{app:software}

In addition to developing the evaluation procedure, we have implemented open-source Python software, \textbf{\textsl{pyIEOE}}, to streamline the evaluation of OPE with our experimental protocol.
This package is built with the intention of being used with \textbf{\textsl{OpenBanditPipeline}} (obp).\footnote{https://github.com/st-tech/zr-obp}

Below, we show the essential codes to conduct an interpretable evaluation of various OPE estimators with our software so that one can grasp the usage of the software easily. Primarily, only four lines of code are sufficient to complete our IEOE procedure in Algorithms~\ref{algo:ieoe} and~\ref{algo:ieoe_real_world} except for some preparations.

\begin{table*}[!htb]
\begin{lstlisting}[title={Code Snippet 1: \textbf{Essential Codes for Interpretable OPE Evaluation}},captionpos=b]
# import InterpretableOPEEvaluator
>>> from pyieoe.evaluator import InterpretableOPEEvaluator

# initialize InterpretableOPEEvaluator class
>>> evaluator = InterpretableOPEEvaluator(
        random_states=np.arange(1000),
        bandit_feedbacks=[bandit_feedback],
        evaluation_policies=[
            (ground_truth_a, action_dist_a),
            (ground_truth_b, action_dist_b),
            ..,
        ],
        ope_estimators=[
            DoublyRobustWithShrinkage(), # DRos
            SelfNormalizedDoublyRobust(), # SNDR
            ..,
        ],
        regression_models=[
            LogisticRegression,
            RandomForest,
            ..,
        ],
        regression_model_hyperparams={
            LogisticRegression: lr_hp,
            RandomForest: rf_hp,
            ..,
        },
        ope_estimator_hyperparams={
            DoublyRobustWithShrinkage.estimator_name: dros_param,
            SelfNormalizedDoublyRobust.estimator_name: sndr_param,
            ..,
        }
    )

# estimate policy values
>>> policy_value = evaluator.estimate_policy_value()

# visualize CDF of squared errors of each OPE estimator
>>> evaluator.visualize_cdf_aggregate()  # plot CDF curves
\end{lstlisting}
\end{table*}

In the following subsections, we explain the procedure including preparations in detail, by showing an example of conducting an interpetable evaluation of OPE estimators on a synthetic bandit dataset.

\subsection{Preparing Dataset and Evaluation Policies}
Before using pyIEOE, we first need to prepare logged bandit feedback data and a set of evaluation policies. Here, each evaluation policy consists of its action distribution and ground-truth policy value. We can conduct this preparation by using the dataset module of obp.

\begin{table*}[!htb]
\begin{lstlisting}[title={Code Snippet 2: \textbf{Preparing Dataset and Evaluation Policies}},captionpos=b]
# import necessary package from obp
>>> from obp.dataset import (
        SyntheticBanditDataset,
        logistic_reward_function,
        linear_behavior_policy
    )
# initialize SyntheticBanditDataset class
>>> dataset = SyntheticBanditDataset(
        n_actions=10,
        dim_context=5,
        reward_type="binary", # "binary" or "continuous"
        reward_function=logistic_reward_function,
        behavior_policy_function=linear_behavior_policy,
        random_state=12345,
    )
# obtain synthetic logged bandit feedback data
>>> bandit_feedback = dataset.obtain_batch_bandit_feedback(n_rounds=10000)
# prepare action distribution and ground truth policy value for each evaluation policy
>>> action_dist_a = #...
>>> ground_truth_a = #...
>>> action_dist_b = #...
>>> ground_truth_b = #...
\end{lstlisting}
\end{table*}

In addition to synthetic dataset, users can utilize multi-class classification data, public real-world data (such as Open Bandit Dataset~\cite{saito2020open}), and their own real-world data to evaluate the robustness of OPE estimators by following preprocessing procedure of obp. Users are also free to define a set of evaluation policies by themselves.

\newpage
\subsection{Defining Hyperparameter Spaces}
After preparing the synthetic data and a set of evaluation policies, we now define hyperparameter spaces of OPE estimators.
Users of the software can define hyperparameter spaces of OPE estimators by themselves as follows.

\begin{table*}[!htb]
\begin{lstlisting}[title={Code Snippet 3: \textbf{Defining Hyperparameter Spaces}},captionpos=b]
# define hyperparameter spaces for ope estimators
>>> lambda_ = {
        "lower": 1e-3,
        "upper": 1e2,
        "log": True,
        "type": float
    }
>>> K = {
    "lower": 1,
    "upper": 5,
    "log": False,
    "type": int
}
>>> dros_param = {"lambda_": lambda_, "K": K}
>>> sndr_param = {"K": K}
# define hyperparameter spaces for regression models
>>> C = {
        "lower": 1e-3,
        "upper": 1e2,
        "log": True,
        "type": float
    }
>>> n_estimators = {
        "lower": 20,
        "upper": 200,
        "log": True,
        "type": int
    }
>>> lr_hp = {"C": C}
>>> rf_hp = {"n_estimators": n_estimators}
\end{lstlisting}
\end{table*}

\newpage
\subsection{Interpretable OPE Evaluation}
Finally, we evaluate OPE estimators in an interpretable manner. 
Our software provides an easy procedure to conduct this evaluation of OPE workflow.

\begin{table*}[!htb]
\begin{lstlisting}[title={Code Snippet 4: \textbf{Interpretable OPE Evaluation}},captionpos=b]
# import InterpretableOPEEvaluator
>>> from pyieoe.evaluator import InterpretableOPEEvaluator
# import other necessary packages
>>> from sklearn.linear_model import LogisticRegression
>>> from sklearn.ensemble import RandomForestClassifier as RandomForest
>>> from obp.ope import DoublyRobustWithShrinkage, SelfNormalizedDoublyRobust

# initialize InterpretableOPEEvaluator class
# define OPE estimators to evaluate
>>> evaluator = InterpretableOPEEvaluator(
        random_states=np.arange(1000),
        bandit_feedbacks=[bandit_feedback],
        evaluation_policies=[
            (ground_truth_a, action_dist_a),
            (ground_truth_b, action_dist_b)
        ],
        ope_estimators=[
            DoublyRobustWithShrinkage(),
            SelfNormalizedDoublyRobust(),
        ],
        regression_models=[
            LogisticRegression,
            RandomForest,
        ],
        regression_model_hyperparams={
            LogisticRegression: lr_hp,
            RandomForest: rf_hp,
        },
        ope_estimator_hyperparams={
            DoublyRobustWithShrinkage.estimator_name: dros_param,
            SelfNormalizedDoublyRobust.estimator_name: sndr_param
        }
    )
# estimate policy values
>>> policy_value = evaluator.estimate_policy_value()
# compute squared errors
se = evaluator.calculate_squared_error()
# compare OPE estimators in an interpretable manner by visualizing CDF of squared errors
>>> evaluator.visualize_cdf_aggregate()  # plot CDF curves

# quantitative analysis by AU-CDF and CVaR
>>> au_cdf = evaluator.calculate_au_cdf_score(threshold=0.004)
>>> print(au_cdf)
{"dr-os": 0.000183.., "sndr": 0.000257..}
>>> cvar = evaluator.calculate_cvar_score(alpha=70)
>>> print(cvar)
{"dr-os": 0.000456.., "sndr": 0.000194..}
\end{lstlisting}
\end{table*}

Users can intuitively evaluate the robustness of the estimators by comparing the CDF of the squared error. 
The quantitative comparison is also possible by calculating some summary scores such as AU-CDF and CVaR. 
In this case, it is easy to figure out that SNDR is more reliable than DRos.